\documentclass[lettersize,journal]{IEEEtran}
\usepackage{amsmath,amsfonts}
\usepackage{algorithm}
\usepackage{algpseudocode}
\usepackage{hyperref}
\usepackage{array}
\usepackage{courier}
\usepackage{xcolor}
\hypersetup{hidelinks}
\usepackage{subfigure}
\usepackage{textcomp,soul}
\usepackage{stfloats}
\usepackage{url}
\usepackage{verbatim}
\usepackage{graphicx}
\usepackage{booktabs}
\usepackage{multirow}
\usepackage{cite}
\usepackage{amsthm}
\usepackage{bm}
\theoremstyle{definition}
\newtheorem{remark}{Remark}

\makeatletter
\pretocmd\@bibitem{\color{black}\csname keycolor#1\endcsname}{}{\fail}
\newcommand\citecolor[1]{\@namedef{keycolor#1}{\color{blue}}}
\makeatother 

\title{UDMC: Unified Decision-Making and Control Framework for Urban Autonomous Driving with Motion Prediction of Traffic Participants}
\author{Haichao Liu, Kai Chen, Yulin Li, Zhenmin Huang, Ming Liu, and Jun Ma
\thanks{Haichao Liu and Kai Chen are with the Robotics and Autonomous Systems Thrust, The Hong Kong University of Science and Technology (Guangzhou), Guangzhou, China (e-mail: hliu369@connect.hkust-gz.edu.cn; kchen916@connect.hkust-gz.edu.cn).}
\thanks{Yulin Li, Zhenmin Huang, Ming Liu, and Jun Ma are with the Department of Electronic and Computer Engineering, The Hong Kong University of Science and Technology, Hong Kong SAR, China (e-mail: yline@connect.ust.hk; zhuangdf@connect.ust.hk; eelium@ust.hk; jun.ma@ust.hk).}
\thanks{This work has been submitted to the IEEE for possible publication.
Copyright may be transferred without notice, after which this version may
no longer be accessible.}}
\begin{document}
\maketitle
\begin{abstract}
Current autonomous driving systems often struggle to balance decision-making and motion control while ensuring safety and traffic rule compliance, especially in complex urban environments. Existing methods may fall short due to separate handling of these functionalities, leading to inefficiencies and safety compromises. To address these challenges, we introduce UDMC, an interpretable and unified Level 4 autonomous driving framework. UDMC integrates decision-making and motion control into a single optimal control problem (OCP), considering the dynamic interactions with surrounding vehicles, pedestrians, road lanes, and traffic signals. By employing innovative potential functions to model traffic participants and regulations, and incorporating a specialized motion prediction module, our framework enhances on-road safety and rule adherence. The integrated design allows for real-time execution of flexible maneuvers suited to diverse driving scenarios. High-fidelity simulations conducted in CARLA exemplify the framework's computational efficiency, robustness, and safety, resulting in superior driving performance when compared against various baseline models. Our open-source project is available at \href{https://github.com/henryhcliu/udmc\_carla.git}{https://github.com/henryhcliu/udmc\_carla.git}.
\end{abstract}

\section{Introduction}

{Advanced driving assistance systems have significantly improved vehicle safety and efficiency. However, the demand for Level 4 (L4) autonomous driving continues to grow, aiming to release drivers from operational tasks}~\cite{claussmann2019review}.
{Typically, autonomous vehicles (AVs) are equipped with multiple sensors to gather information about the surrounding traffic conditions~\cite{prakash2021multi,chen2023milestones}. To compensate for the range and accuracy deficiencies of physical perception, the vehicle to everything (V2X) technique has been extensively studied~\cite{deng2019cooperative, wang2019survey,lee2022design}, which can be utilized as a complement or even a replacement for physical sensors.}
\begin{figure}[t]
\centering
\subfigure[Merge into the roundabout]{
\includegraphics[width=0.465\linewidth]{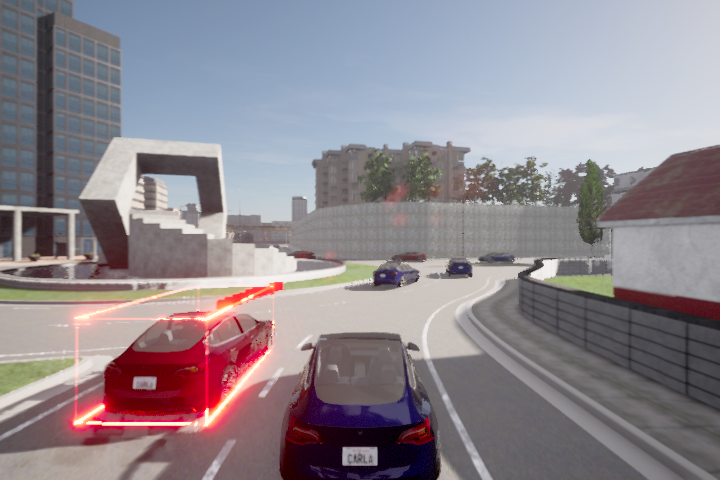} \label{fig:roundabout_show1}
}
\subfigure[Drive in the roundabout]{
\includegraphics[width=0.465\linewidth]{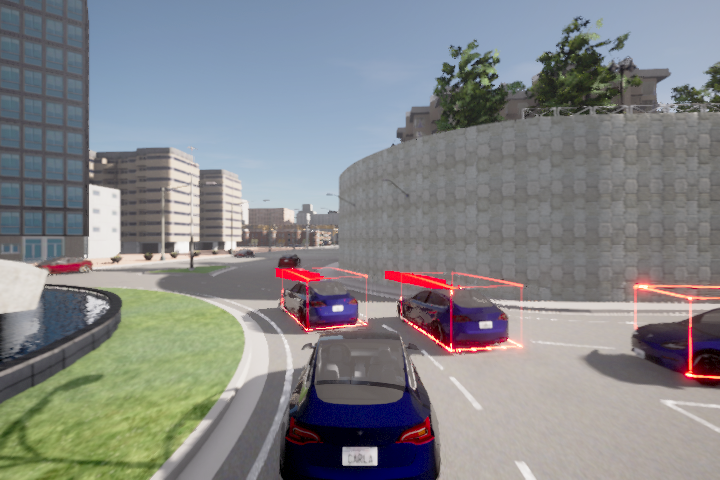} \label{fig:roundabout_show2}
}
\caption{{An intelligent vehicle with the proposed driving framework is driving in a roundabout with congested traffic conditions. (a) The vehicle is merging into the roundabout when {a surrounding vehicle} is blocking another lane. (b) The vehicle is driving in the inner lane when there are two {surrounding vehicles} in front and one {surrounding vehicle} attempting to merge in.}}
\label{fig:roundabout_show}
\end{figure}
{Upon perceiving environmental data, trajectory prediction of the traffic participants is necessary.}
{The existing literature extensively explores various prediction methods including both model-based~\cite{eiras2021two} and data-driven methods~\cite{liu2024incremental,wang2023sa}.}

{Benefiting from the advancements in perception and prediction, the development of effective decision-making and control schemes has attracted considerable research attention, especially in various complicated urban driving scenarios illustrated in Fig.~\ref{fig:roundabout_show}.} Intuitively, hierarchical frameworks are intensively investigated, which generally split the decision-making, motion planning, and control processes into consecutive modules\cite{zhu2023hierarchical,yao2024hierarchical}. Motion planning frameworks that cover both multi-lane and single-lane scenarios are proposed in~\cite{fan2018baidu}, which employs a high-level strategy for lane changing and a low-level inner-lane strategy for iterative path and speed optimization. In~\cite{zhang2017finite}, a rule-based finite state machine (FSM) behavior planner is utilized for highway driving scenarios, incorporating complex modality switch conditions. {Responsibility-Sensitive Safety (RSS) provides a safe distance model for the longitudinal and lateral directions, but it is conservative and traffic efficiency is not appropriately considered~\cite{shalev2017formal}.} 
Another popular hierarchical method achieving high-level path planning is graph searcher~\cite{qi2022hierarchical}, which utilizes both heuristic and partial motion planning for locating rough paths in spatiotemporal space. Reinforcement learning is also successfully utilized to make robust and complex lane-changing and lane-merging decisions under perception uncertainty~\cite{he2022robust,hwang2022autonomous}. Typically, the above decision-making or behavior-planning methods need a lower-level motion control module to execute the control commands, such as PID~\cite{zhang2017finite}, Iterative LQR~\cite{chen2019autonomous,ma2022alternating,ma2022local,huang2023decentralized,huang2023parallel}, and Model Predictive Control (MPC)~\cite{xu2019design}. 
{The above hierarchical frameworks demand significant human design or a large amount of driving data, yet it remains challenging to encompass all possible driving scenarios due to the long-tail effect~\cite{wu2023integrated}. Furthermore, although the low-level control module is time efficient, facilitating real-time performance is still challenging since all tasks from decision to control must be completed sequentially within a restricted time frame. Moreover, integrating different units into a hierarchical structure could be cumbersome and rather time-consuming.}

{The rapid advancement of integrated methods for autonomous driving is notable. These methods primarily fall into two categories. Learning-based approaches, such as InterFuser utilizing transformers~\cite{shao2023safety}, and the two-stage learning process proposed in \cite{chen2020learning}, are intensively studied. 
To enhance user safety, a spatial-temporal feature learning scheme called ST-P3 is proposed\cite{hu2022st}. Afterward, VAD is presented to reduce the computing overhead and implement instance-level scenario construction~\cite{jiang2023vad}.
Reinforcement learning techniques, as reported in~\cite{zhang2021enabling, cai2022dq, li2022decision}, have also demonstrated efficacy in urban driving. However, these methods encounter challenges in adapting to varied driving scenarios~\cite{omeiza2022}.
In contrast, optimization-based methods ensure interpretability~\cite{guan2022integrated} by formulating an integrated OCP to obtain optimal strategies. In \cite{sahin2020autonomous, eiras2021two}, mixed-integer programming (MIP) is utilized for decision-making and obtaining discretized decision variables. In addition, game-based approaches like trajectory game model~\cite{zhu2023game} are proposed to deal with interactions between agents. However, solving MIP and game-based problems is notably time-consuming~\cite{hang2020integrated}. While linear-quadratic-Gaussian approximations can reduce the computational burden~\cite{schwarting2021stochastic}, dealing with complex driving conditions remains challenging.}

{To enhance computational efficiency, artificial potential field (APF) methods~\cite{wang2019obstacle, zhou2022autonomous} integrate various forms of potential functions (PFs) for simultaneous decision-making and control tasks, offering flexibility and efficiency. Safety Force Field (SFF) maps world perception into constraints on control to directly prevent collisions~\cite{nister2019safety}. Notably, MPC schemes with simple PFs demonstrate successful collision avoidance in straight-line driving scenarios~\cite{rasekhipour2016potential}. Further, FSM is incorporated into the APF-MPC scheme to assist lane change maneuvers~\cite{canale2022autonomous}, but this strategy limits generalizability across driving scenarios. For curvy roads, the Frenet frame is commonly used to execute autonomous driving~\cite{lim2021hybrid, lim2018hierarchical}. Yet, the Frenet frame cannot model the actual vehicle dynamics, and the vehicle shape is distorted. In \cite{li2022autonomous}, a trajectory planner in the Cartesian frame on curvy roads is proposed, which focuses on simplified one-lane scenarios, restricting broader urban applications. 
Furthermore, traffic light regulations, a critical aspect, are often overlooked. An APF-based risk field model addresses traffic light constraints on vehicular movement~\cite{hua2022modeling, wang2022driving, wang2024a}, primarily for longitudinal vehicle planning. Therefore, designing specific PFs to adhere to traffic rules in diverse urban scenarios remains a challenge~\cite{liu2023integrated}. }
\begin{figure}[t]
    \centering
    \includegraphics[width=1\linewidth]{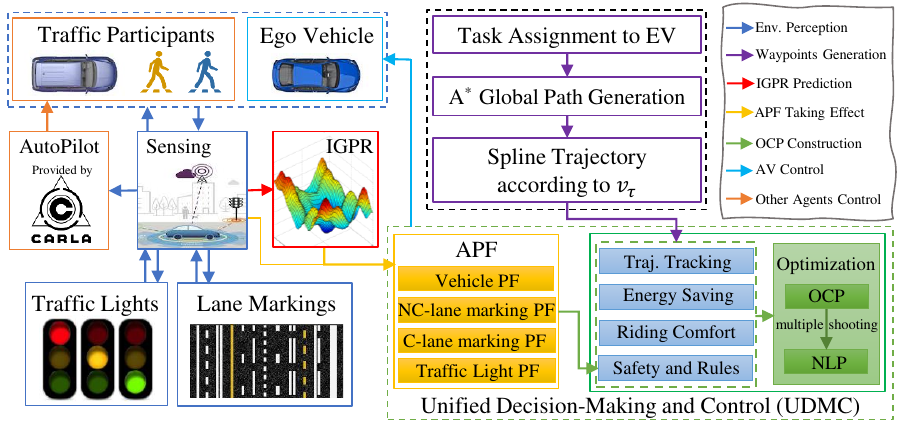}
    \caption{The proposed unified decision-making and control {(UDMC)} framework for urban autonomous driving. The arrows with different colors represent different kinds of information transmission routes.}
    \label{fig:udmc_framework}
\end{figure}

{In order to enhance computational efficiency and adaptability to diverse urban driving scenarios, while prioritizing driving safety and adherence to traffic rules, this paper introduces a unified decision-making and control (UDMC) scheme as depicted in Fig.~\ref{fig:udmc_framework}.}
With motion prediction of {surrounding vehicles} based on Interpolation-based Gaussian Process Regression (IGPR) and feature characterization of {surrounding vehicles} and traffic rules as incorporated in the APF, our UDMC framework leverages an appropriate OCP formulation to address various urban driving scenarios. 
The key contributions of this work are as follows.
\begin{itemize}
    \item {We propose an interpretable L4 autonomous driving framework for unified decision-making and motion control, which facilitates safe, robust, and traffic rule-complied driving in complex urban scenarios.}
    \item {Innovative PFs are developed to model various traffic participants and regulations. By integrating them into the OCP, our framework enables real-time generation of flexible maneuvers for the AV.}
    \item {Our proposed IGPR is effectively incorporated for rapid and reliable motion prediction of surrounding vehicles and pedestrians, providing essential guidance for rational decision-making and safe motion control.}
    \item The UDMC framework is deployed in CARLA, and the high-fidelity urban driving tests with tons of curvy roads clearly demonstrate the computational efficiency, robustness, and safety guarantee attained with our UDMC scheme. {This project is open-source to push forward} the development of autonomous driving frameworks.
\end{itemize}

The rest of this paper is organized as follows. Section II presents the preliminaries of this work. Section III illustrates the general pipeline of the proposed autonomous driving framework. In Section IV, UDMC is proposed with an appropriate OCP formulation. Next, Section V details the comparison and ablation simulations to evaluate the proposed UDMC. Finally, Section VI concludes this work.
\section{Preliminaries}
\subsection{Notation}
We use the bold lowercase letter and uppercase letter to represent a vector $\bm x \in \mathbb{R}^n$ and a matrix $\bm X \in \mathbb{R}^{n\times m}$, respectively. The vector formed by the $i$th to $j$th elements of a vector $\bm x$ is denoted as $\bm x_{[i:j]}$. The $i$th row and $j$th column of a matrix $\bm X$ is denoted by $[\bm X]_{i,\cdot}$ and $[\bm X]_{\cdot,j}$, respectively. Also, $\text{diag}\{a_1, a_2, \ldots, a_n\}$ represents a diagonal matrix with $a_i, i \in \{1,2,\ldots, n\}$ as diagonal entries.
The set of non-negative integers is represented as $\mathbb N$.
{$[\, \cdot\, ]^i_\tau$ represents the state vector of the $i$th object at the time step $\tau$. 
It is important to note that various objects possess their unique state vectors. For instance, a vehicle is characterized by a state vector consisting of position, velocity, and heading angle. Similarly, the state vector of a road lane center includes position and orientation, while the state vector of a traffic light object is defined by the position of the corresponding stop line.} 
If the state is in the {AV} frame, we additionally impose a superscript $[\, \cdot\, ]^\text{ev}$ in the notation. For simplicity, a matrix $[\bm s^1 ,\bm  s^2 , \ldots,\bm  s^n]$ is denoted as $[\bm s^i], i \in \{1,2,\ldots, n\}$. We use $\mathcal{N}(\bm \mu, \bm \Sigma)$ to describe a Gaussian distribution with the mean vector $\bm \mu$ and the covariance matrix $\bm \Sigma$. The squared weighted 2-norm $\bm x^T \bm M \bm x$ is simplified as $\| \bm x \|^2_{\bm M}$. 

\subsection{Vehicle Dynamics Model}\label{subsec:2-1VehicleDynamicsModel}
In this paper, a nonlinear vehicle dynamics model proposed in~\cite{ge2021numerically} is used. This model exhibits good numerical stability and high fidelity in the scene of frequent acceleration and deceleration, as typically observed in urban driving tasks.
The state vector of this model is $\bm x = [p_x, p_y, \varphi, v_x, v_y, \omega]^T\in \mathbb R^6$, and we denote $\bm p = [p_x, p_y]^T$ as the vehicle positions in $x$-axis and $y$-axis, $\bm v = [v_{x}, v_y]^T$ is a vector that contains the longitudinal and lateral velocities. Besides, $\varphi$ is the heading angle, and $\omega$ is the vehicle's yaw rate. The control input vector of the system is $\bm u = [a,\delta]^T\in \mathbb R^2$, where $a$ and $\delta$ represent the acceleration and steering angle, respectively. We utilize the discrete dynamics model $\bm x_{\tau+1}=f(\bm x_\tau, \bm u_\tau)$ with the time step $T_s$ derived from backward Euler method\cite{ge2021numerically}:
\begin{equation}
\label{eq:bicycle_dynamics}
\begin{split}
    \bm x_{\tau+1}
    =\left[\begin{array}{c}
    p_{x,\tau}+T_s\left(v_{x,\tau} \cos \varphi_{\tau}-v_{y,\tau} \sin \varphi_{\tau}\right) \\
    p_{y,\tau}+T_s\left(v_{y,\tau} \cos \varphi_{\tau}+v_{x,\tau} \sin \varphi_{\tau}\right) \\
    \varphi_{\tau}+T_s \omega_{\tau} \\
    v_{x,\tau}+T_s a_{\tau} \\
    \frac{m v_{x,\tau} v_{y,\tau}+T_s L_k \omega_{\tau}-T_s k_f \delta_{\tau} v_{x,\tau}-T_s m v_{x,\tau}^2 \omega_{\tau}}{m v_{x,\tau}-T_s\left(k_f+k_r\right)} \\
    \frac{I_z v_{x,\tau} \omega_{\tau}+T_s L_k v_{y,\tau}-T_s l_f k_f \delta_{\tau} v_{x,\tau}}{I_z v_{x,\tau}-T_s\left(l_f^2 k_f+l_r^2 k_r\right)}
    \end{array}\right]
\end{split},
\end{equation}
where $m$ is the vehicle mass, $l_f$ and $l_r$ are the distance from the mass center to the front and rear axle, $k_f$ and $k_r$ are the cornering stiffness of the front and rear wheels, and $I_z$ is the inertia polar moment. Note that in (\ref{eq:bicycle_dynamics}), we define $L_k=l_f k_f - l_r k_r$ for simplicity.
\subsection{Gaussian Process Regression for Motion Prediction}\label{subsec:2-3GPR}
A Gaussian Process~(GP) is a non-parametric data modeling method that defines a collection of random variables with joint Gaussian distribution. Given a set of $m$ training data
\begin{equation}\label{eq:dataset}
\begin{aligned}
\mathcal{H}=\{\bm{Y} & =\left[\bm{y}_1, \bm y_2 \ldots, \bm{y}_m\right] \in \mathbb{R}^{n_y \times m}, \\
\bm{Z} & \left.=\left[\bm{z}_1, \bm{z}_2, \ldots, \bm{z}_m\right] \in \mathbb{R}^{n_z \times m}\right\}
\end{aligned}
\end{equation}
with input $\bm Z$ and output $\bm Y$, a GP can be expressed as $\mathcal{GP}(m(\bm x),k(\bm x, \bm x'))$, where $m(\bm x)$ is the mean function, and $k(\bm x, \bm x')$ is the covariance function of $\bm x$ and $\bm x'$. This function defines the prior properties of the GP for inference. In GPR, the prediction of a set of continuous quantities $\bm y_\tau \in \mathbb R^{n_y}$ is inferred from a collection of inputs $\bm z_\tau \in \mathbb R^{n_z}$ by the function $g(\bm z_\tau):\mathbb R^{n_z}\to \mathbb R^{n_y}$, which is
\begin{equation}\label{eq:GPR}
    \bm y_\tau = g(\bm z_\tau) + \bm \epsilon_\tau,
\end{equation}
where $\bm \epsilon_\tau \sim \mathcal{N}(0,\Sigma^{\bm \epsilon})$ is i.i.d. Gaussian noise with variance $\Sigma^{\epsilon}=\text{diag}\{\sigma_1, \sigma_2, \ldots, \sigma_{n_y}\}\in \mathbb R^{n_y\times n_y}$, and
\begin{equation}\label{eq:g(z)}
    g(\bm z_\tau)\sim \mathcal{N}(\bm \mu^{y}(\bm z_\tau), \bm \Sigma^{y}(\bm z_\tau))
\end{equation}
is a Gaussian distribution with the mean $\bm \mu ^{y}(\bm z) = [\mu^1(\bm z),\mu^2(\bm z), \ldots, \mu^{n_y}]^T\in \mathbb R^{n_y}$, and the covariance $\bm \Sigma^y(\bm z) = \text{diag}\{\Sigma^1(\bm z),\Sigma^2(\bm z), \ldots, \Sigma^{n_y}(\bm z)\}\in \mathbb R^{n_y\times n_y}$ in every dimension of the predicted states. For each dimension, the posterior distribution at the new state $z_\tau$ is Gaussian with
\begin{align}\label{eq:mu(z)}
\mu^i(\bm{z}) & =\bm{\phi}_{\bm{z Z}}^i\left(\bm{\Phi}_{\bm{Z Z}}^i+\bm{I} \sigma_i^2\right)^{-1}[\bm{Y}]_{\cdot, i}, \\
\begin{split}\label{eq:Sigma(z)}
\Sigma^i(\bm{z}) & =\phi_{\bm{z z}}^i-\bm{\phi}_{\bm{z Z}}^i\left(\bm{\Phi}_{\bm{Z Z}}^i+\bm{I} \sigma_i^2\right)^{-1} \bm{\phi}_{\bm{Z} \bm{z}}^i, 
\end{split}
\end{align}
where $\mu^i(\bm z)$ and $\Sigma^i(\bm{z})$ are the mean and variance of the $i$th element of $g(\bm z)$, $\bm Y\in \mathbb R^{m \times n_y}$ is the output of the training data. $\bm{\Phi}_{\bm{Z Z}}^i$ is a symmetric positive semi-definite Gram matrix, where
\begin{equation}\label{eq:gramMatCal}
\bm {\Phi_{ZZ}}^i=\left[\begin{array}{cccc}
\phi_{11}^i& \phi_{12}^i & \ldots & \phi_{1n_d}^i \\
\phi_{21}^i& \phi_{22}^i & \ldots & \phi_{2n_d}^i \\
\ldots & \ldots & \ldots & \ldots \\
\phi_{n1}^i& \phi_{n2}^i & \ldots & \phi_{n_d n_d}^i \\
\end{array}\right].
\end{equation}
Note that $n_d$ is the number of data points, $\phi_{jk}^i$ in the matrix is the covariance between the $j$th and $k$th data points for the $i$th element of $\bm y$ calculated by a kernel function. In this paper, we use the Gaussian radial basis function (RBF) as the kernel, which is defined as
\begin{equation}
    \phi_{jk}^i = \beta_i^2 \exp{(\gamma_i \left\|\bm z_i - \bm z_j \right\|^2)},
\end{equation}
where $\beta_i\in [0,1]$ and $\gamma_i = -1/2l^2$ are applied to adjust the shape of the kernel. Similarly, $\phi_{\bm{z z}}^i = \beta_i^2 \exp{(\gamma_i \left\|\bm z - \bm z \right\|^2)}=\beta_i^2$, and $[\bm{\phi}_{\bm{z Z}}^i]_j = \beta_i^2 \exp{(\gamma_i \left\|\bm z_j - \bm z \right\|^2)}$. Hence, after the training process, Gram matrices $\{\bm{\Phi_{ZZ}^i}, i=1,2,\ldots,m\}$ are obtained from the training data $\mathcal{H}$ in (\ref{eq:dataset}). Finally, we can infer the result $\bm y_\tau$ from input data $\bm z_\tau$ at time step $\tau$ from the environment by (\ref{eq:GPR}) and its affiliate functions (\ref{eq:g(z)})-(\ref{eq:Sigma(z)}).

\section{Pipeline of the Proposed Autonomous Driving Framework}\label{sec:pipeline}
{In Fig. \ref{fig:udmc_framework}, we introduce a comprehensive decision-making and control framework leveraging motion prediction for traffic participants. This autonomous driving framework comprises three key modules: (a) global path planning and smoothing, (b) surrounding information collection and motion prediction, and (c) unified decision-making and control guided by APF.}

\subsection{Global Path Planning and Smoothing}\label{subsec:3-1globalPP}
This section outlines our method for global path planning and trajectory smoothing. We employ a two-step process: initial path creation via the A$^*$ algorithm, followed by spline interpolation for drivable and smooth trajectories.

\subsubsection{Preliminary Global Path Generation} {The foundation of our path planning strategy is to create a connectivity graph that represents possible paths on the road. We start by generating waypoints, which are equidistant points positioned along the centerline of each lane within the OpenDRIVE map. These waypoints serve as nodes in the connectivity graph. To establish the graph's edges, we link each waypoint to its immediate predecessor and successor within the same lane. We also create connections to corresponding waypoints in the neighboring left and right lanes to account for lane changes.}

{Given any starting point $\bm x^s = [p^{s}_x, p^{s}_y, \varphi^s]^T$ and a target destination $\bm x^t = [p^{t}_x, p^{t}_y, \varphi^t]^T$, we employ the A$^*$ algorithm on this graph to identify the shortest available path between these two points. The advantage of this approach lies in its efficiency and the ability to enhance the continuity of the path, as the search is conducted over a simplified representation of the actual road network.}

\subsubsection{Path Smoothing Using Spline Interpolation} {Once the A$^*$ algorithm has determined the preliminary set of global reference waypoints, we proceed to refine this path to ensure that it is smooth and can be followed easily by the vehicle. This refinement is achieved through spline interpolation, generating local reference waypoints that are feasible for the vehicle to track over time.
Let ${[x_i, y_i]^T}, {i\in \{0,1,\cdots,n\}}$, be the set of $n+1$ knots, where $x_i$ is the $i$th knot's position along the x-axis, and $y_i$ is the corresponding value of the function to interpolate. The cubic spline $S(x)$ is a function such that in each interval $[x_i, x_{i+1}]$, the spline $S(x)$ is represented by a cubic polynomial $y_i=P_i(x)$:
\begin{equation} 
S(x) = 
\begin{cases} 
P_0(x) & \text{for } x_0 \leq x < x_1,\\ P_1(x) & \text{for } x_1 \leq x < x_2,\\ ~\vdots \\ P_{n-1}(x) & \text{for } x_{n-1} \leq x \leq x_n. \end{cases} 
\end{equation}
Each polynomial $P_i(x)$ can be written as: \begin{equation} P_i(x) = a_i + b_i(x - x_i) + c_i(x - x_i)^2 + d_i(x - x_i)^3, \end{equation} where $a_i$, $b_i$, $c_i$, and $d_i$ are the coefficients that need to be determined for each subinterval.
The spline interpolation process involves solving for these coefficients such that the spline is differentiable and the derivatives are continuous at the knots.}
{Spline interpolation is chosen for its mathematical properties, which guarantee the smoothness of the resulting path. By fitting a spline curve through the global reference waypoints, we obtain a trajectory that is not only continuous but also differentiable.
Subsequently, the algorithm generates a set of tracking points, where the distance between adjacent points is dictated by the interval time and the target velocity.}
\begin{remark}
    {Compared with the original built-in \texttt{autopilot} in CARLA, our presented global planning and smoothing method not only adjusts the distance between adjacent waypoints based on the vehicle's reference speed but also refines these waypoints dynamically considering the current speed of the vehicle. This adaptability allows for a more precise alignment with the features of subsequent waypoints at each time step, to better facilitate MPC in the driving system.}
\end{remark}

\subsection{Surrounding Information Collection and Motion Prediction}\label{subsec:SurrInfoCollec}
Since this work focuses on the design of our UDMC scheme {committed to effectively avoiding obstacles under the premise of traffic rule compliance, we don't assign a specific environment model/sensing solution. Yet, the semantic information required by the UDMC is clarified as follows:}

To avoid obstacles, it is necessary to obtain the surrounding vehicles' state vector $\bm x^i = [p^{i}_x, p^{i}_y, \varphi^s, v^{i}_x, v^{i}_y] \in \mathbb R^5$ within the range $r_p$ around the {AV}. {Moreover, to predict the motion of the surrounding vehicles for AV's better driving decision, their historical states within the last $N$ time steps, i.e., $\bm x^i_\tau,\tau\in\{-N,\cdots,-1\}$ are required as well. Lastly, for} the vehicles to obey the traffic rules, it is necessary to obtain the state vector of the current line center $\bm p^i = [p_{x}^i, p_{y}^i, \varphi^i]^T$, the type of lane markings $tp \in TP = \{\texttt{crossable}, \texttt{non-crossable}\}$ on both sides where the {AV} is currently located, as well as the current state of the traffic lights $tl \in TL = \{\texttt{red}, \texttt{green}\}$ near the {AV}. 
{In terms of lane markings,} there are various types on the road, such as white broken lines, white solid lines, yellow double solid lines, etc. 

{Note that this information collection module is agnostic to specific sensing technologies. Hence,  the UDMC scheme can seamlessly integrate with various perception approaches (such as camera, LiDAR, and so on), thereby enhancing its applicability across different vehicle platforms.}
\subsection{Unified Decision-Making and Control with APF Guidance}
In dynamic and complicated urban environments, it is crucial to efficiently formulate and solve decision and control problems for autonomous driving to ensure their safety and improve efficiency. The optimization objective, which is typically formulated in a receding horizon fashion, aims to track the waypoints obtained from spline interpolation (Section~\ref{subsec:3-1globalPP}), maintain target velocity to enhance driving efficiency, and smoother the control input to make passengers more comfortable while avoiding collisions and traffic rule violations. Constraints encompass the vehicle's dynamics model, state and control command boundaries, predicted future motion for {surrounding vehicle} avoidance, and adherence to traffic rules such as lane markings and traffic lights. Notably, traffic rule constraints are incorporated into potential fields, directing the AV away from potential hazards. The unified decision and control approach enables the vehicle to swiftly respond to unforeseen circumstances, leveraging comprehensive surrounding perception.
\section{Unified Decision Making and Control Scheme for Urban Autonomous Driving}
With various PFs, the proposed UDMC scheme can implicitly generate decision-making commands (such as overtaking, cutting in, or following) in a receding horizon fashion. It receives the environmental perception and predicts the surrounding traffic objects' future path, makes implicit and continuous decisions, and outputs the acceleration and steering angle of the {AV}.
\begin{figure*}
\centering
\subfigure[$F_\text{CR}$ and $F_\text{NR}$]{
\includegraphics[width=0.23\linewidth]{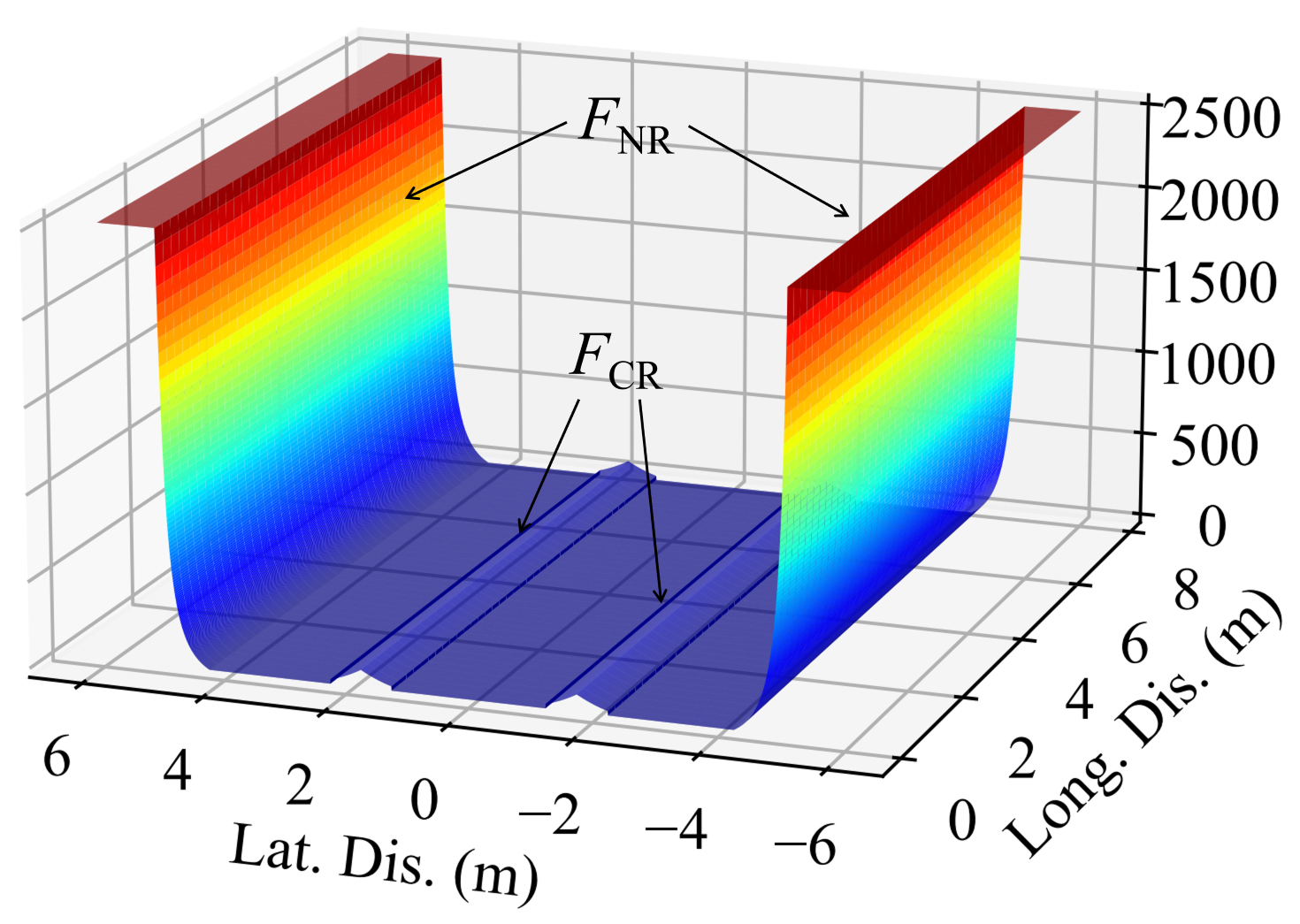} \label{APF_laneMarkings}
}
\subfigure[$F_\text{TL}$]{
\includegraphics[width=0.23\linewidth]{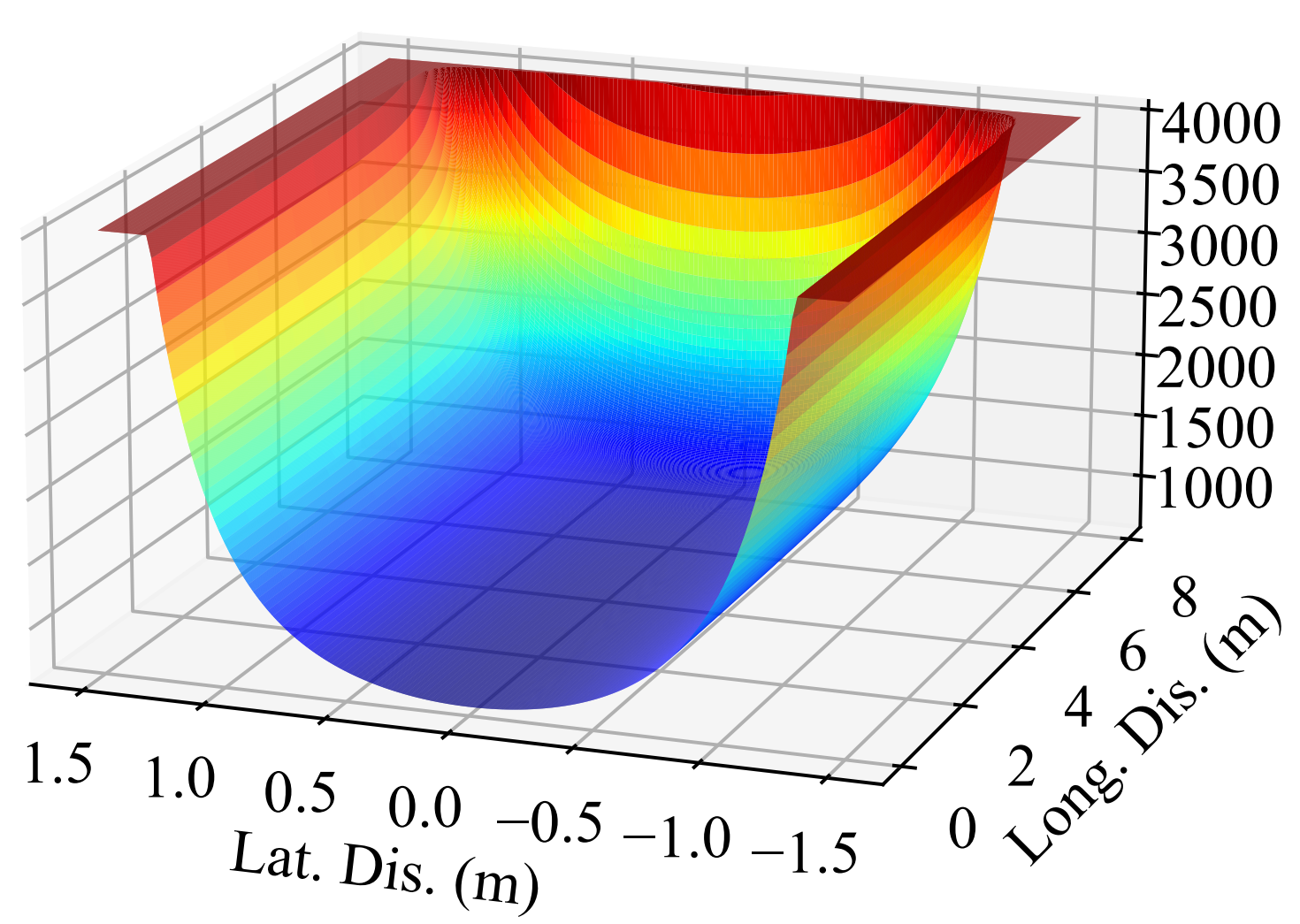} \label{APF_traffic}
}
\subfigure[{$F_\text{VC}$ and $F_\text{PD}$}]{
\includegraphics[width=0.23\linewidth]{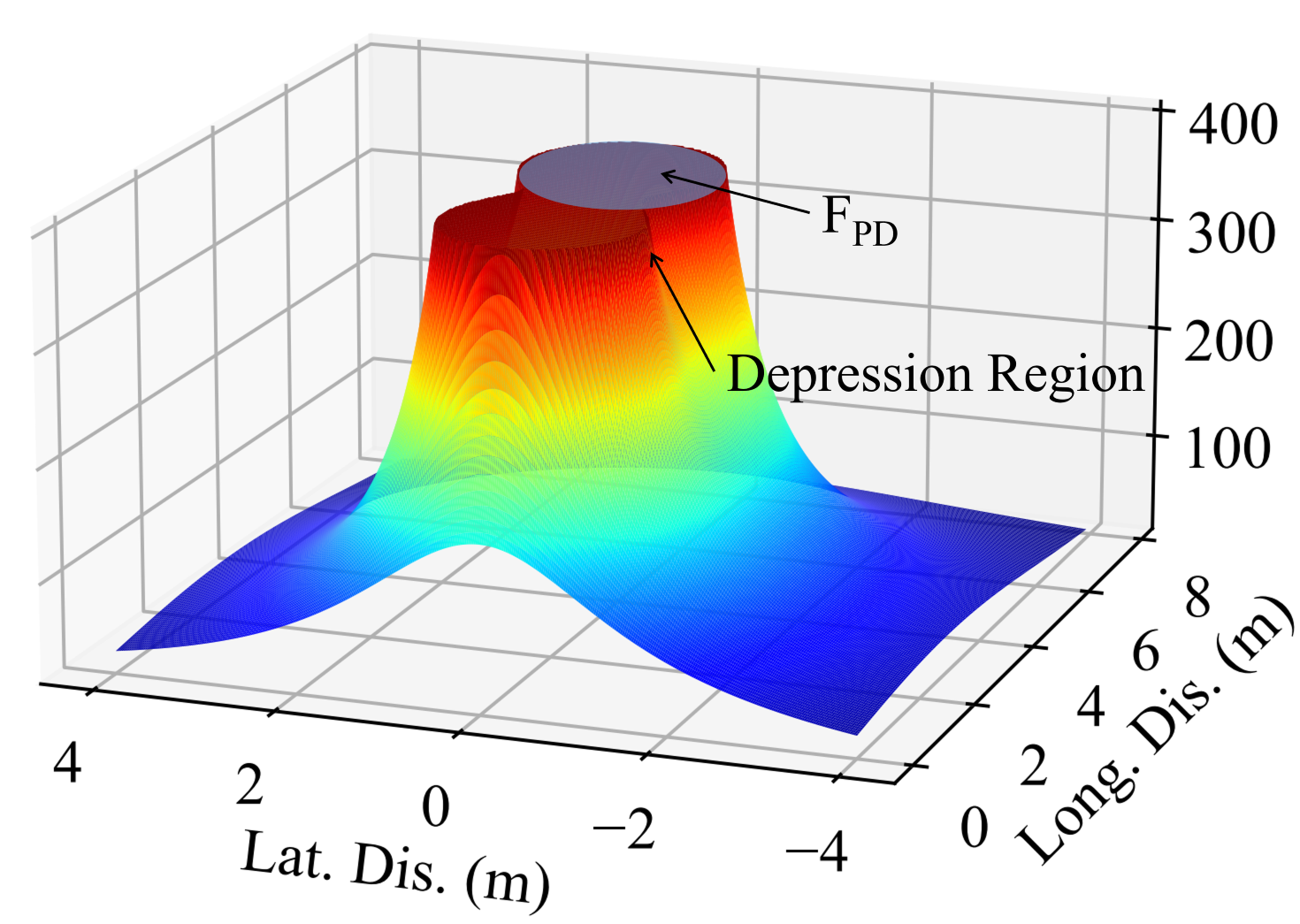} \label{APF_vehicles}
}
\subfigure[$F_\text{VE}$]{
\includegraphics[width=0.23\linewidth]{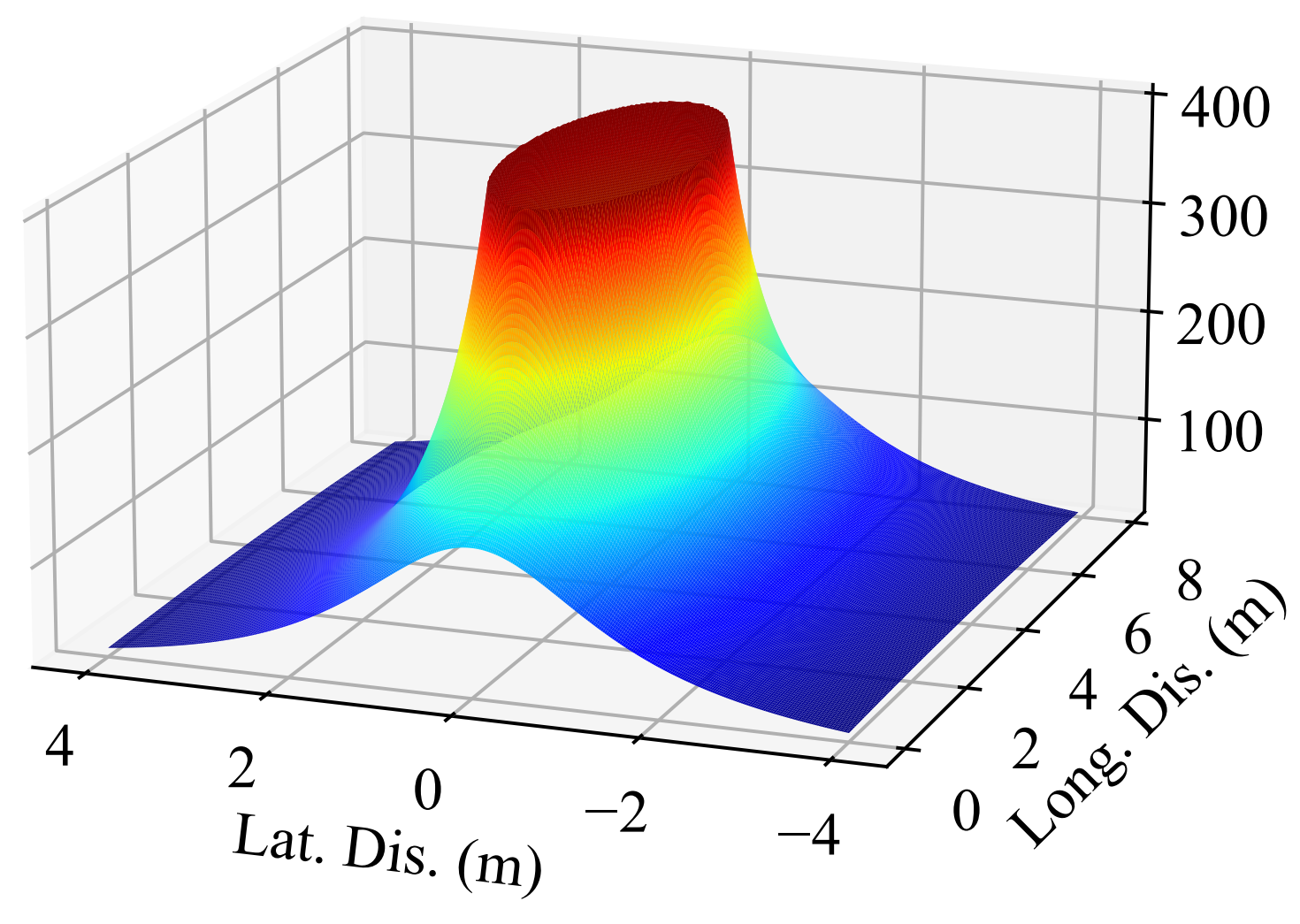} \label{APF_vehicles_ellip}
}
\caption{{The designed potential functions for traversable and non-traversable lane markings, {surrounding vehicles}, and red traffic light signals. (a) $F_\text{CR}$ and $F_\text{NR}$ for crossable and non-crossable lanes. The potential field is summarized by two traversable and two non-traversable lane markings constituting a three-lane road structure. The lateral position of non-traversable markers and traversable markers are [5.25, -5.25] m and [1.75, -1.75] m, respectively. (b) $F_\text{TL}$ for traffic control signals. The PF of the red traffic light, where the parameters of the PF are set as $p_y^\text{ev} = 8.0$ m. (c) $F_\text{VC}$ for vehicles with circles wrapping. The PF is obtained by wrapping a vehicle using two circles, where the vehicle's position is set as $[p^k_x, p^k_y]=[3.0,0]$ m. (d) $F_\text{VE}$ for vehicles with an ellipsoid wrapping. The PF is obtained by wrapping a vehicle using an ellipsoid to construct the second version of vehicle PF, where $[p^k_x, p^k_y]=[4.0,0]$ m.}}\label{fig:APF_rendering}
\end{figure*}
\subsection{Potential Functions Design for Traffic Objects}\label{subsec:4-1APFDesign}
The PFs are designed for {surrounding vehicles}, traffic lights, and two kinds of road lines. The 3D shape of the APF generated by the designed PFs is rendered in Fig.~\ref{fig:APF_rendering}, from which one can understand how the PFs affect the decisions of the proposed UDMC. Typically, the APF is centered on the obstacle or the goal, which emits a repulsive or attractive force field to the environment so that the system is driven to the destination without collisions. In this paper, the repulsive PF for {traffic participants, including {surrounding vehicles} and pedestrians}, the repulsive PF for non-crossable lane markings, and the attractive PF for the centerline of lanes are suitably designed. Unlike the traditional APF method where the target position is set as a gravitational field~\cite{weerakoon2015artificial}, this paper employs novel penalty terms for collision avoidance and traffic rules compliance. Therefore, the potential field in the unified decision and control module is characterized as
\begin{equation}
\label{eq:sum_PFs}
\begin{aligned}
    {F(\bm p^\text{env}, \bm x)} & = \sum_i F_{\text{NR}}^i + \sum_j F_{\text{CR}}^j {+ F_{\text{TTC}}} \\
    &+ \sum_k F_{\text{V}}^k + {\sum_q F_{\text{PD}}}+ F_{\text{TL}},
\end{aligned}
\end{equation}
where $i\in \{0,1,2\}$, $j\in \{0,1,2\}$, $k\in \mathbb N$, and $q\in \mathbb N$ denote the indices of non-crossable lane markings, crossable lane markings, {{surrounding vehicles}, and pedestrians, respectively.
In addition, $\bm p^\text{env}$ represents all related environmental state vectors of the traffic system. This includes the state vectors of {surrounding vehicles}, pedestrians, various road lanes, and traffic lights.}
Notice that, to simplify the PF design, we classify various types of lane markings on the road into two categories as {described in Section~\ref{subsec:SurrInfoCollec}.}

The lateral distance from the {AV} to the lane marking is essential to determine the PF force. We define the road centerline state vector as $\bm p^i = [p_{x}^i, p_{y}^i, \varphi^i]^T$, where the first two items $p_{x}^i$ and $p_{y}^i$ represent the position of the current road centerline in the global Cartesian frame, and the last item $\varphi^{i}$ is the tangential angle between the current road and the global $x$-axis. Therefore, the lateral distance of the lane marking to {AV} is given as
\begin{align} \label{cordinate transform of rd}
s_{\text{R}}(\bm x, \bm p^i)=\left\{
\begin{aligned} 
&\left({p}_y^\text{ev}(\bm p^i) + \frac{w_R}{2}\right) - y^\text{ev}(\bm x, \bm p^i) &\text{left}\\ 
&y^\text{ev}(\bm x, \bm p^i) - \left(p_y^\text{ev}(\bm p^i) - \frac{w_R}{2}\right)
&\text{right}
\end{aligned}
\right.  
\end{align}
where ``left'' and ``right'' indicate whether the lane marking is on the left or right side of the vehicle, $w_R$ is the lane width, $p_y^\text{ev}$ and $y^\text{ev}$ are the lateral positions of the current lane centerline and the {AV} in the heading angle-paralleled (HAP) frame. In the HAP, the orientation of the road is paralleled with the $x$-axis, and the origin is the same as the world frame. Thus, the above lateral positions in the HAP frame are
\begin{align}
&p_y^{\text{ev}}(\bm p^i) = p_{x}^i \sin(\varphi^i) + p_{y}^i\cos(\varphi^i),\\
    \begin{split}
        &y^{\text{ev}}(\bm x,\bm p^i) = p_{x} \sin(\varphi^i) + p_{y}\cos(\varphi^i).
    \end{split}
\end{align}
Given credit to the above distance calculation, the PFs for lane markings can be effectively expressed as follows.
\subsubsection{Non-Crossable Lane Markings}\label{subsubsec:NR PF}
 To comply with traffic rules and ensure driving safety, traversing a solid lane marking is prohibited. Especially when the lane marking is next to curbstones or a sidewalk, crossing results in potential danger. Therefore, when the distance between the {AV} and the surrounding non-crossable lane markings decreases to a certain extent, the designed PF $F_\text{NR}$ should give a repulsive force that increases rapidly with decreasing distance to urge the {AV} to steer away immediately. Based on the inverse proportional function and power function, we design the following PF for the $i$th non-crossable lane marking:
\begin{align} \label{eq:NR_Def}
F_{\text{NR}}^i=\left\{
\begin{aligned} 
&m_s&s_{\text{R}_i}\leq0.1\\ 
&\frac{a_\text{NR}}{s_{\text{R}}(\bm x, \bm p^i)^{b_\text{NR}}} - e_s& 0.1<s_{\text{R}_i}<1.5\\
&0& s_{\text{R}_i} \geq 1.5
\end{aligned}
\right.
\end{align}
where $a_\text{NR}$ and $b_\text{NR}$ are the intensity and shape parameters that control the strength of the repulsive force and the effective range of the PF. The smooth parameters are designed as
$$e_s = \frac{a_\text{NR}}{1.5^{b_\text{NR}}},\ m_s = \frac{a_\text{NR}}{0.1^{b_\text{NR}}}-e_s,$$ {which ensures the continuity of the PF value when $s_{\text{R}_i}$ transitions on both sides of 0.1\,m and prevents the occurrence of abnormally large values.} The 3D shape of the designed non-crossable lane marking's PF is illustrated on both sides of Fig. \ref{APF_laneMarkings}, which illustrates how the repulsive force urges the {AV} to drive away from the road boundaries.
\subsubsection{Crossable Lane Markings}
In the process of overtaking and lane changing, it is inevitable to traverse the broken lane marking on the road. 
When the reference lane of the {AV} is blocked from ahead by other traffic participants, the {AV} needs to switch to an adjacent lane and keep driving until the reference lane is free.
When the {AV} is driving on a lane that is not the reference, designing a PF that keeps the {AV} driving along the centerline is necessary, otherwise the {AV} will be attracted to the side of the current lane due to the attractive force of the reference lane. Moreover, the $F_\text{CR}$ is beneficial to quick and decisive lane change behavior so that the {AV} does not keep on the crossable lane marking anymore after the lane changing decision.
Combine the above requirements, the PF of the $j$th crossable lane marking is designed as follows:
\begin{align} \label{CR-PF}
F_{\text{CR}}^j=\left\{
\begin{aligned} 
&a_\text{CR} (s_{\text{R}}(\bm x, \bm p^j)-b_\text{CR})^2& s_{\text{R}_j}<b_\text{CR}\\
&0& s_{\text{R}_j} \geq b_\text{CR}
\end{aligned}
\right.
\end{align}
where $a_\text{CR}$ is the intensity parameter and $b_\text{CR}$ is the effective range of the repulsive force from the lane marking. Such PF repulses the {AV} away from the lane marking and forces it to stay around the lane center. The 3D shape of the crossable lane marking PF is shown in the middle area of Fig. \ref{APF_laneMarkings}, which illustrates how the $F_\text{CR}$ attracts the {AV} to drive near the centerline.

For the {surrounding vehicle}'s PF $F_\text{V}$ design, we examine two methods: double-circle wrapped $F_\text{VC}$ and ellipsoid-wrapped $F_\text{VE}$.

\subsubsection{Surrounding Vehicles}
Here, we approximate each on-road vehicle with two identical circles aligned in the longitudinal direction that cover the entire vehicle.
In order to avoid {surrounding vehicles} smoothly and quickly, we design the following PF:
\begin{equation}\label{Vehicle_PF}
    {F_{\text{VC}_k} = \sum_{m=1}^2\sum_{o=1}^2 \frac{a_\text{V}}{((p_x^m-p_{x}^{k,o})^2+(p_y^m-p_{y}^{k,o})^2)^{b_\text{V}}},}
\end{equation}
where $a_\text{V}$ and $b_\text{V}$ are the intensity and shape parameters. They control the strength and the effective range of this PF, respectively. As two circles represent the vehicle, we must calculate the circles' positions and sum up their forces to formulate the integrated PF. The position vector $[p_{x}^{k,o},p_{y}^{k,o}]^T$ denotes the $o$th circle position of the $k$th {surrounding vehicle}, and we have
\begin{equation}\label{CalcCircleCenters}
\left[
\begin{array}{c}
p_{x}^{k,o} \\
p_{y}^{k,o}
\end{array}
\right]=
\left[
\begin{array}{cc}
1& 0 \\
0& 1
\end{array}
\right]
\left[
\begin{array}{c}
p^k_x \\
p^k_y
\end{array}
\right]
\pm
\left[
\begin{array}{c}
\cos{\varphi^k}\\
\sin{\varphi^k}
\end{array}
\right] r_\text{V},
\end{equation}
where the common radius of the circles is $r_\text{V}=1.4$ and $o \in \{1,2\}$ indicates whether the circle is front or rear. {Similarly, the position vector of the AV $[p_x^m,p_y^m]^T,m\in\{1,2\}$ can be calculated by (\ref{CalcCircleCenters}).}
Fig. \ref{APF_vehicles} illustrates this potential field of double-circle-wrapped vehicle in front of the {AV}.

{Alternatively}, in this work, we only use an ellipsoid to represent a surrounding vehicle.
Leveraging the notations above, {the corresponding PF} $F_{\text{VE}_k}$ is
\begin{equation}\label{Vehicle_PF_ellip}
    {F_{\text{VE}_k} = \sum_{m=1}^2\frac{a_\text{V} (r_a r_b)^2}{(r_b^2(p_x^m-p_{x}^{k})^2+r_a^2(p_y^m-p_{y}^{k})^2)^{b_\text{V}}},}
\end{equation}
where $r_a$ and $r_b$ are the semi-major and semi-minor axis of the ellipsoid wrapping an {surrounding vehicle}. Fig.~\ref{APF_vehicles_ellip} demonstrates this potential field of an ellipsoid-wrapped vehicle in front of the {AV}.
Notice that $F_\text{VC}$ and $F_\text{VE}$ share the same notation of $F_\text{V}$.
\begin{remark}\label{remark:doubleCircles}
    {Compared with (\ref{Vehicle_PF}), this PF reduces the complexity of the OCP by reducing the number of items for collision avoidance and consequently simplifies the original OCP. Another superiority of the ellipse-based expression is that without the depression region of the double-circle-based expression shown in Fig.~\ref{APF_vehicles}, the OCP is easier to solve with fewer non-convex elements.
    }
\end{remark}
{To capture safety margins w.r.t. relative velocity and mitigate the risks associated with on-road interactions, we exert an extra PF inspired by time to collision (TTC). TTC is calculated by dividing the distance between two vehicles by their relative velocity, i.e.,}
\begin{equation}
    {TTC_k^2=\frac{(p_x-p_x^k)^2+(p_y-p_y^k)^2}{(v_x-v_x^k)^2}\in[0,+\infty).}
\end{equation}
{Assign $s=-TTC^2\in (-\infty,0]$, according to the physical meaning of TTC, larger $s$ is supposed to give higher repulsive force to the AV. Hence, we impose the following item along with the PF for the leading vehicle, namely $F_\text{TTC}$, to the APF:}
\begin{equation}
    {F_\text{TTC} = a_T \left(e^{b_T\left(s + t_{\text{alarm}}^2\right)} - 1\right)+F_{\text{V}_l},}
\end{equation}
{where $a_\text{T}$ and $b_\text{T}$ are the intensity and shape parameters like other designed PFs, while the alarming threshold is denoted as $t_{\text{alarm}}$, and $F_{\text{V}_l}$ is the PF for the \textit{leading vehicle} with the same formula defined in (\ref{Vehicle_PF_ellip}). As for the utility of the PF, on the one hand, if the real-time TTC exceeds $t_{\text{alarm}}$, $F_\text{TTC}$ increases sharply from 0 to large values. On the other hand, if TTC is large, $F_\text{TTC}$ remains negative with a small gradient. As a result, the AV keeps a safe margin from the leading vehicle and effectively prevents rear-end collisions.}

\subsubsection{Pedestrians}
{Without loss of generality, pedestrians are crucial traffic participants in urban driving scenarios. We use one circle centered at the pedestrian's head from the bird-eye view to avoid collisions with pedestrians, which is similar to the expression of $F_\text{VC}$. Specifically, the PF of pedestrian $F_{\text{PD}}$ is designed as:
\begin{equation}\label{Pedestrian_PF}
    F_{\text{PD}} = \frac{a_\text{PD}}{((p_x-p_{x}^{k})^2+(p_y-p_{y}^{k})^2)^{b_\text{PD}}},
\end{equation}
where $a_\text{PD}$ and $b_\text{PD}$ are the intensity and shape parameters, while $[p_x^k,p_y^k]^T$ is the position of the $k$th pedestrian near the {AV}. As illustrated in Fig.~3(c), the contour of $F_{\text{PD}}$ closely resembles one-half of $F_\text{VC}$.
}
\subsubsection{Traffic Lights}
In addition to obeying the rules of lane markings, vehicles must also obey the command of traffic lights. Traditionally, when a stop signal is issued by the traffic light ahead, the {AV} is simply commanded to brake in front of the stop line, but this strategy neglects the comfort of passengers. Therefore, this paper proposes a novel PF for traffic lights, which can restrict the lateral and longitudinal positions of the {AV} simultaneously and guide the vehicle to stop smoothly. The PF applies repulsive forces to the {AV} in three directions without changing its original reference trajectory. In particular, the PF is designed as
\begin{equation}
\label{TL_equation}
    F_\text{TL} = c_\text{TL}\left(\frac{a_{\text{TL}_1}}{d_{x}^\text{ev}}+\frac{a_{\text{TL}_2}}{d_{y,l}^\text{ev}}+\frac{a_{\text{TL}_2}}{d_{y,r}^\text{ev}}\right),
\end{equation}
where $c_\text{TL}$ is an indicator of the state of the traffic light, which is equal to 0 if the traffic light is $green$ and equal to 1 otherwise.
Besides, $a_{\text{TL}_1}$ and $a_{\text{TL}_2}$ are intensity parameters in the longitudinal and lateral orientations, respectively, and ${d_{x}^\text{ev}}$ is the longitudinal distance from the {AV} to the stop line $\bm p^q$. Moreover, ${d_{y,l}^\text{ev}}$, $d_{y,r}^\text{ev}$ are the distance of the {AV} to the left and right lane markings, respectively, which can be calculated by (\ref{cordinate transform of rd}) as well. Fig.~\ref{APF_traffic} shows the PFs that enclose the {AV} from three directions.

As elaborated, the above PFs are compatible with almost all urban driving scenarios and can readily describe various traffic situations effectively and precisely.
\begin{figure}[t]
\centering
\subfigure[{VPF for unmarked road section}]{
\includegraphics[width=0.90\linewidth]{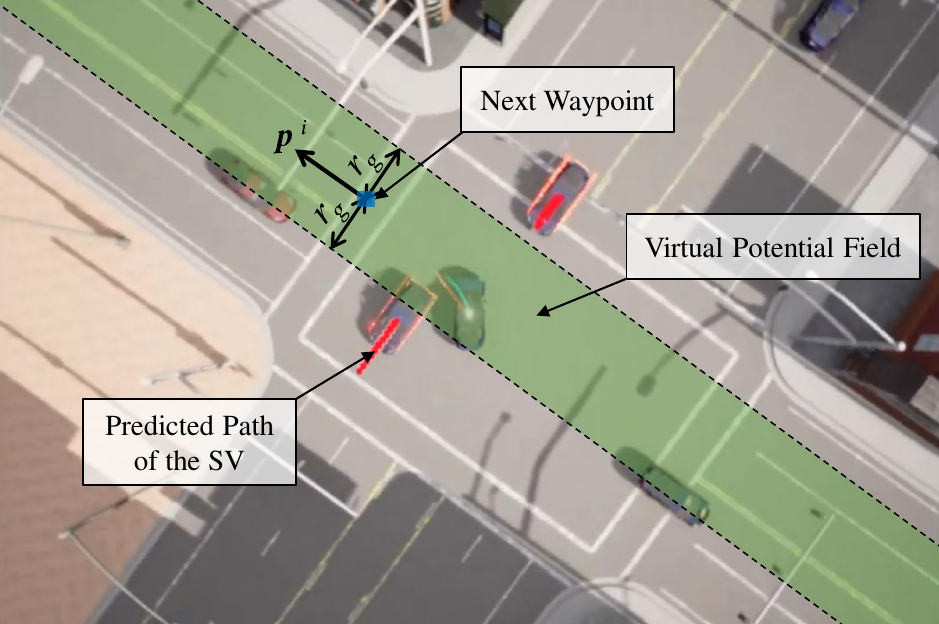} \label{fig:VPF_intersection}
}
\subfigure[{VPF for turning left behavior}]{
\includegraphics[width=0.90\linewidth]{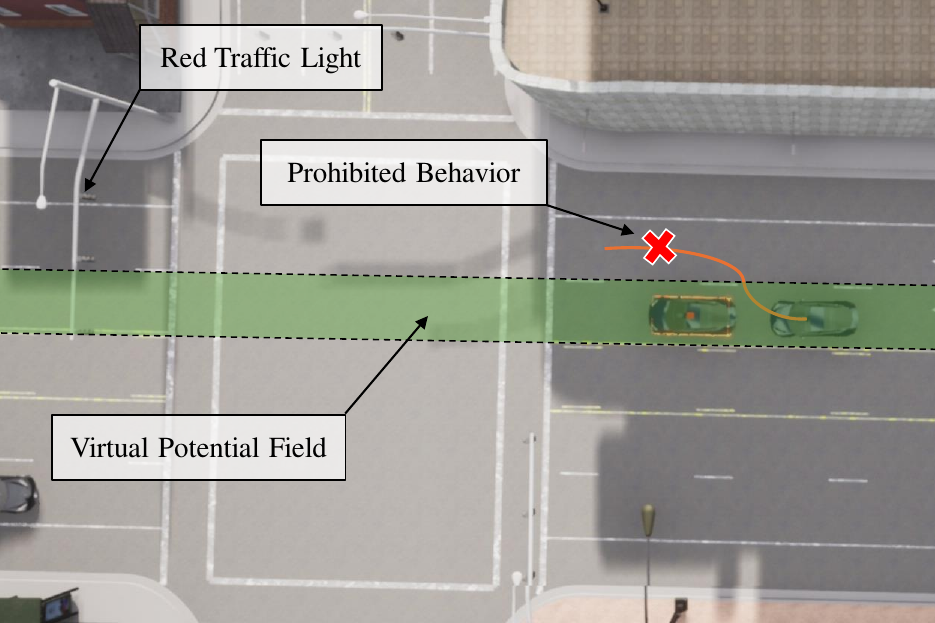} \label{fig:VPF_turn_left}
}
\caption{{Illustration of the effectiveness of the proposed VPF. (a) The VPF exists in an intersection, where the vehicle is crossing the intersection with the protection of the proposed VPF to avoid diverging too much from the target destination when {surrounding vehicles} are nearby. (b) The VPF exists before turning left, where the vehicle is waiting behind the leading vehicle, and driving to another road lane is prohibited by traffic rules.}}\label{fig:VPF_utility}\label{fig:gpf_show}
\end{figure}
\subsubsection{Virtual Potential Field in Complex Traffic Conditions}
To handle complex traffic conditions such as un-signalized intersection turning, we propose the Virtual Potential Field (VPF) to guarantee further driving safety and traffic rule compliance. First, we propose the VPF for the intersection area, as shown by the green field {in Fig.~\ref{fig:VPF_intersection}.} This figure illustrates a situation where there is an {surrounding vehicle} in front of the {AV}, which is prone to diverge too much to get a small objective value if no PF is guiding its way. Hence, the proposed VPF gives virtual guidance to the {AV} to drive to the line-marked areas. {Specifically, the VPF contains two virtual non-crossable lane markings defined by (\ref{eq:NR_Def}), which have a larger distance 
\begin{equation}
    r_g=r_\text{lane}+r_\text{offset},
\end{equation}
compared with standard road lane PFs.} Here, $r_\text{lane}=w_R/2$ is half of the road lane width, and $r_\text{offset}$ is a relaxation factor.
\begin{remark}
    {We relax the condition of the VPF boundaries to create a slightly larger free region for the {AV} to escape from the {surrounding vehicles}. Only if the center of the {AV} does not exceed the corresponding real road lanes, the {AV} would be able to adjust its lateral position with the guidance of $F_\text{NR}$, which is also the reason we set $r_\text{lane}=w_R/2$.}
\end{remark}
Second, we design the VPF for the turning area {as shown in Fig.~\ref{fig:VPF_turn_left}.} When the {AV} is waiting to turn left or right in front of the intersection, it is supposed to stay in the correct lane even if there is no vehicle in the adjacent lanes. Our strategy is to detect the three waypoints ahead of the {AV} to determine whether a turning behavior is approaching. If a turning behavior is detected at time step $\tau$, and the distance $s_R(\bm x_\tau, \bm p_\tau)$ between the {AV} and the next target waypoint is less than $r_\text{lane}$, we set the opposite traffic lane PF to be non-crossable. If 
\begin{equation}
    s_R(\bm x_\tau, \bm p_\tau)>r_\text{lane},
\end{equation}
a higher {trajectory-following penalty} is exerted to urge the {AV} to get back in the correct lane. According to the above analysis, the {AV} is urged to stay in the initially planned lane by global path planning, which complies with the correct driving lane in intersections.

{
To sum up, the presented VPF ensures the {AV}'s safety and adherence to traffic regulations by keeping it within a secure zone, regardless of the number of {surrounding vehicles} trying to repel it from the intersections or turning lanes.
}

\subsection{Implementation of Surrounding Vehicles' Potential Field with Motion Prediction}\label{subsec:SVPredictIGPR}
To deploy the PFs designed in Section~\ref{subsec:4-1APFDesign}, the future motion of {surrounding vehicles} must be predicted. Considering the unified and light-weighted nature of UDMC, we adopt the GPR mechanism to the above prediction task. One of the advantages of GPR for prediction is its interpretability, which is essential to the proposed optimization-based UDMC scheme with a safety guarantee.
As introduced in Section\,\ref{subsec:2-3GPR}, GPR is a non-parametric Bayesian approach to regression, which provides not only the predicted values but also the uncertainty estimation of the predictions. In our task, the GPR model is applied to predict the 10 future states $\bm p_\tau = [x_\tau, y_\tau, \varphi_\tau]^T\in\,\mathbb R^3, \tau \in \{1,2,\ldots,10\}$ of the {surrounding vehicles} based on the 15 history states $\bm p_\tau = [x_\tau, y_\tau, \varphi_\tau, v_{x,\tau}, v_{y,\tau}]^T\in\,\mathbb R^5, \tau \in \{-14,-13,\ldots,0\}$.

A set of training inputs $\bm{Z}$ and corresponding outputs $\bm{Y}$ are collected in the simulator with $n_s$ {surrounding vehicles} running on urban roads around the whole town map. Hence, the positions of the vehicles have a considerable extensive range from around -300\,m to 300\,m. We formulate the dataset as follows for training:
\begin{equation}\label{eq:dataset_sv}
\begin{aligned}
\mathcal{H}=\{\bm{Y} & =\left[\bm{y}_{1}, \bm y_2, \ldots, \bm{y}_{m}\right] \in \mathbb{R}^{30 \times m}, \\
\bm{Z} & \left.=\left[\bm{z}_{1}, \bm z_2, \ldots, \bm{z}_{m}\right] \in \mathbb{R}^{75 \times m}\right\},
\end{aligned}
\end{equation}
where $\bm y_i = \left[ \bm p_1, \bm p_2, \ldots, \bm p_{10}  \right]\in \mathbb R^{30}, i = \{1,2,\ldots,m\}$, and $\bm z_i = \left[ \bm p_{-14}, \bm p_{-13}, \ldots, \bm p_0  \right]\in \mathbb R^{75}, i = \{1,2,\ldots,m\}$. By organizing the {surrounding vehicle} path data using the above method, we connect the history states and the future states by timeline, so it contains time information inherently. Notice that the independent variable of this GPR is not time steps used in the existing literature~\cite{yu2023sparse}. Hence, we reformulated the extrapolation prediction problem into its counterpart.

Another improvement to the GPR lies in the training process. We pre-process the data in $\mathcal{H}$ by position centering to speed up the convergence, making the GPR prediction more accurate. For each data pair of 15 history states and 10 future ones, we subtract the first state to ensure every piece of data record starts from the center of the map, which eliminates the effect of position on the GPR training process by
\begin{equation}
    [\bm p_\tau]_j = [\bm p_\tau]_j - [\bm p_{-14}]_j, j = \{1,2\}.
\end{equation}
The above centralization operation denoises the dataset for more accurate prediction performance in a smaller data value range. Based on the above improvements, we propose the IGPR to take advantage of GPR in interpolation prediction. It's worth pointing out that IGPR works within the interpolation way from the following two aspects: First, the input of IGPR is history states rather than the timeline, and this eliminates the extrapolation nature of prediction tasks. Second, the centralization operation gathers all the data pairs into a finite position range, which guarantees that the newly collected input is also from the origin of the transformed coordinate system. Hence, it is an interpolation from the other two pieces of data with marginal larger and smaller $\varphi$ values.

When the data is ready, the IGPR model can be trained to learn the kernel's hyperparameters described in Section~\ref{subsec:2-3GPR}, such as the length scale $l$ and the noise level $\beta^2$. The training consists of finding the optimal values of these hyperparameters by maximizing the likelihood of the observed data. In our implementation, we use a maximum likelihood estimation with a specified number of restarts for the optimizer to find the optimal hyperparameters. Once the IGPR model is trained, it can predict the future vehicle state vectors $\bm y$ for any given history {surrounding vehicle} state vector $\bm z$. The prediction is provided by (\ref{eq:GPR}) as a Gaussian distribution with a mean vector $\boldsymbol{\mu}(\bm z)$ and a covariance matrix $\boldsymbol{\Sigma}(\bm z)$, representing the predicted future state vector and the uncertainty of the predictions, respectively.

As the future states of the {surrounding vehicles} are predicted, we sum the APF $F_\tau(\bm p^\text{env}, \bm x_\tau)$ at the time steps $\tau\in \{1,2,\ldots,N\}$ as part of the objective function
\begin{align}
\label{eq:sum_PFs_time}
    F(\bm P, \bm X) & = \sum_{\tau=1}^N\left(\sum_{i=1}^I F_{\text{NR},\tau}^i \left(\bm p^i_\tau, \bm x_\tau\right) +F_{\text{TL},\tau}\left(\bm p^q_\tau, \bm x_\tau\right)\right)\notag\\
    +\sum_{\tau=1}^N&\left(\sum_{k=1}^K F_{\text{V},\tau}^k\left(\bm p^k_\tau, \bm x_\tau\right) + \sum_{j=1}^J F_{\text{CR},\tau}^j \left(\bm p^j_\tau, \bm x_\tau\right)\right), 
\end{align}
where three dimensional matrix $\bm P = \{\bm p_1^\text{env},\bm p_2^\text{env}, \ldots, \bm p_N^\text{env}\}\in \mathbb R^{M\times N}$, and $\bm p^\text{env}_\tau = \{[\bm p_\tau^i], [\bm p_\tau^j],[\bm p_\tau^k],[\bm p_\tau^q]\}\in \mathbb R^{3\times(I+J+K+Q)}$. $M$ denotes the dimension of $\bm p_\tau^\text{env}$. Here, leveraging the prediction result of the proposed IGPR, the $k$th {surrounding vehicle}'s future state $[\bm p_\tau^k]$ at time step $\tau$ is $\bm y_{[\tau\times3-2: \tau\times3]}^k$. Hence, we can construct a potential field gathering the above PFs by~(\ref{eq:sum_PFs_time}).
\subsection{Optimal Control Problem Formulation}
This section formulates the decision-making and control scheme, which leverages the receding horizon strategy to formulate an OCP. 
The cost function is formulated as 
\begin{align}
\label{basic_cost}
\begin{aligned}
    J(\bm X, \bm U, \bm P) &= \sum_{\tau=1}^N \| \boldsymbol x_{\text{ref},\tau}- \boldsymbol x_\tau\| ^2_{\bm{Q}} + \sum_{\tau=1}^N \|\boldsymbol u_{\tau}\|^2 _{\bm{R}} \\
    & +\sum_{\tau=2}^N \|\bm u_{\tau}-\bm u_{\tau-1}\|^2 _{\mathbf{R}_d}+F(\bm P, \bm X),
\end{aligned}
\end{align}
where $N$ is the length of the receding horizon, $X = \{\bm x_1,\bm x_2, \ldots, \bm x_N\}\in \mathbb R^{6\times N}$, $U = \{\bm u_1,\bm u_2, \ldots, \bm u_N\}\in \mathbb R^{2\times N}$,  $\bm x_{\text{ref},\tau}\in \mathbb{R}^{6}$ is the reference trajectory point generated by spline interpolation based on the global path at the time step $\tau$. In (\ref{basic_cost}), $\mathbf{Q}\in \mathbb{R}^{6\times 6}$, $\mathbf{R}\in \mathbb{R}^{2\times 2}$, $\mathbf{R}_d\in \mathbb{R}^{2\times 2}$ are positive semi-definite diagonal weighting matrices for penalizing reference tracking, energy saving, and passenger comfort, respectively. $F(\bm p_\text{env}, \bm x)$ is the potential field generated according to the environmental information, as described in (\ref{eq:sum_PFs}).

Therefore, the OCP is constructed as:
\begin{equation}
\label{NMPCOptProb}
\begin{array}{ll}
\underset{{\boldsymbol{x_\tau}}, {\boldsymbol{u_\tau}}}{\min} & J(\bm X, \bm U, \bm P)\\
\text { s.t.} & \boldsymbol x_{\tau+1}=f(\boldsymbol x_\tau,\boldsymbol u_\tau), \forall \tau \in \{1,2,...,N\} \\
& -\boldsymbol u_\text{min} \preceq \boldsymbol u_\tau \preceq \boldsymbol u_\text{max}, \forall \tau \in \{1,2,...,N\} \\
& -\boldsymbol x_\text{min} \preceq \boldsymbol x_\tau \preceq \boldsymbol x_\text{max}, \forall \tau \in \{1,2,...,N\},
\end{array}
\end{equation}
where the first constraint represents the vehicle dynamics (\ref{eq:bicycle_dynamics}), while the second and third constraints indicate the restriction on the control inputs and system state variables from the physical properties of the selected vehicle model.
\begin{remark}
    {
    The term ``unified'' refers to the fact that our proposed framework integrates decision-making and motion control functionalities into a single OCP. The unified framework enables the vehicle to make decisions and execute maneuvers simultaneously, resulting in a more efficient and streamlined driving experience.
    }
\end{remark}

Afterward, Problem (\ref{NMPCOptProb}) is reformulated to a nonlinear programming (NLP) problem, and the direct multiple shooting method can be used to solve it numerically, in which we treat the states at shooting nodes as decision variables and improve convergence by lifting the NLP problem to a higher dimension.

\section{Simulation Results}\label{sec:simulation}
{In this section, we begin by optimizing vehicle dynamics parameters using SLSQP. Subsequently, we rigorously assess our UDMC framework across diverse urban driving scenarios in CARLA. Comparative evaluations against three baselines and a general ablation study are conducted, followed by deployment and comparison on the CARLA Town05 Benchmark for broader scenario coverage.}
\begin{figure*}[htbp]
\centering
\subfigure[ML-ACC: $t=10.60$\,s]{
\includegraphics[width=0.23\linewidth]{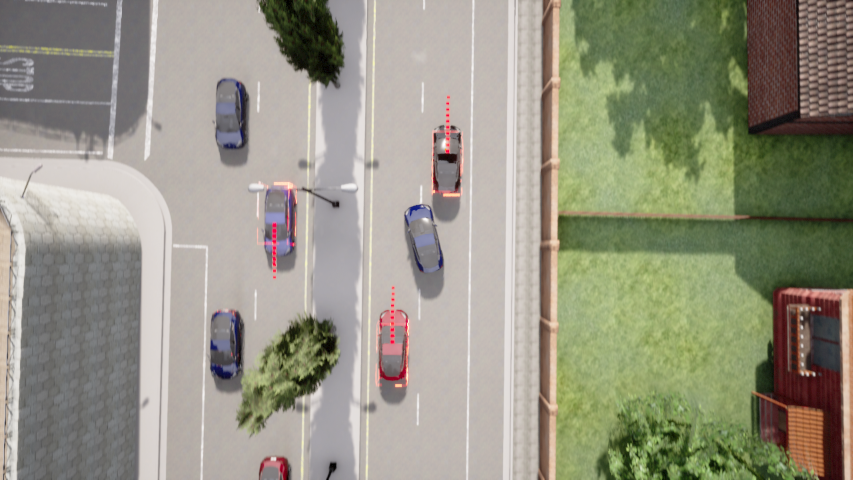}
\label{fig:MLACC1}
}
\subfigure[ML-ACC: $t=12.05$\,s]{
\includegraphics[width=0.23\linewidth]{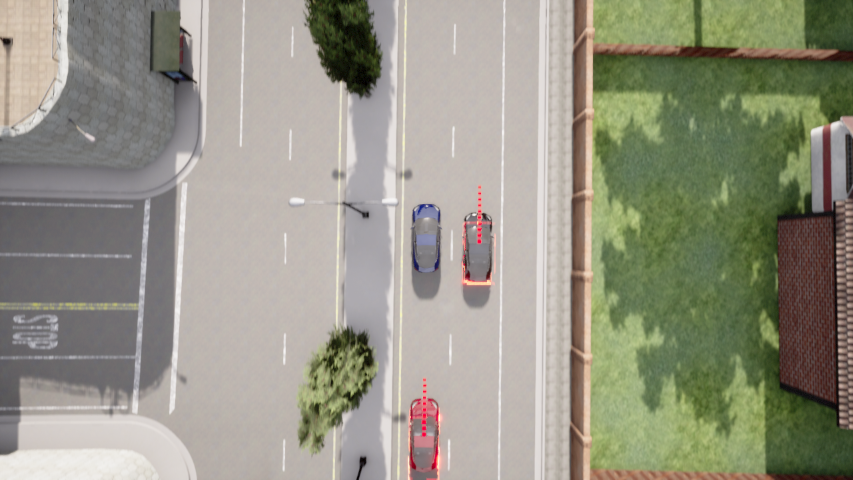}\label{fig:MLACC2}
}
\subfigure[ML-ACC: $t=14.00$\,s]{\includegraphics[width=0.23\linewidth]{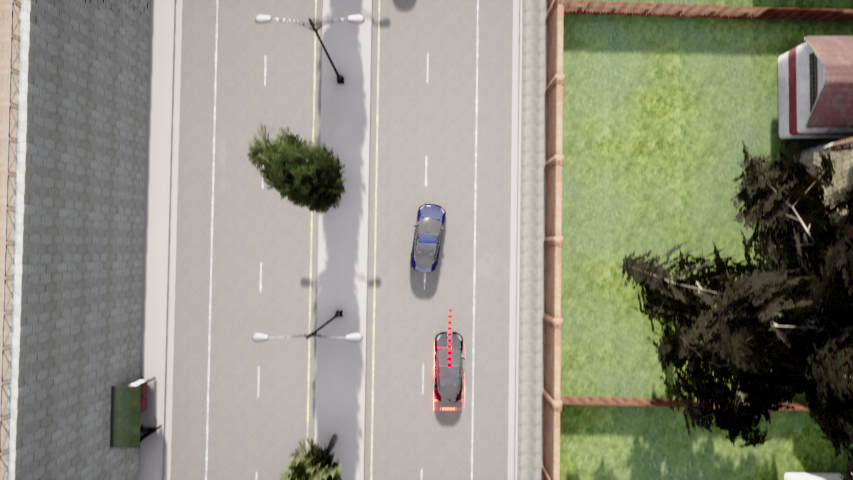}                  
\label{fig:MLACC3}}
\subfigure[ML-ACC: $t=15.90$\,s]{
\includegraphics[width=0.23\linewidth]{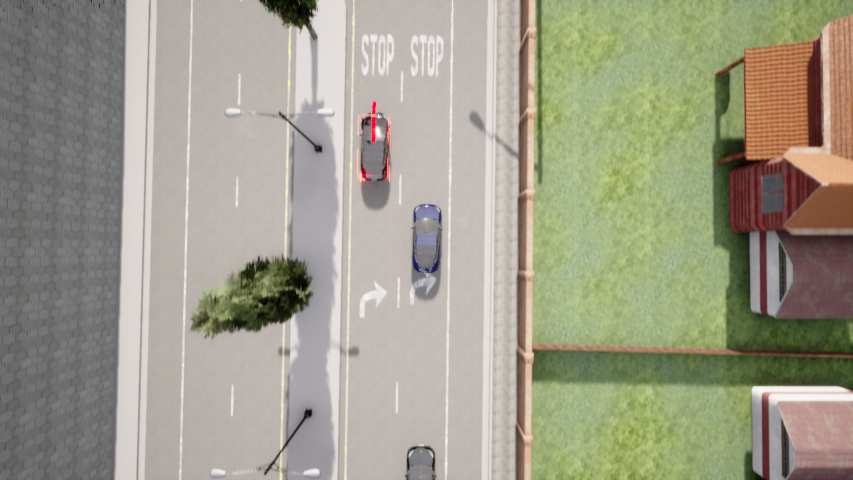}
\label{fig:MLACC4}
}
\subfigure[Roundabout: $t=1.10$\,s]{
\includegraphics[width=0.23\linewidth]{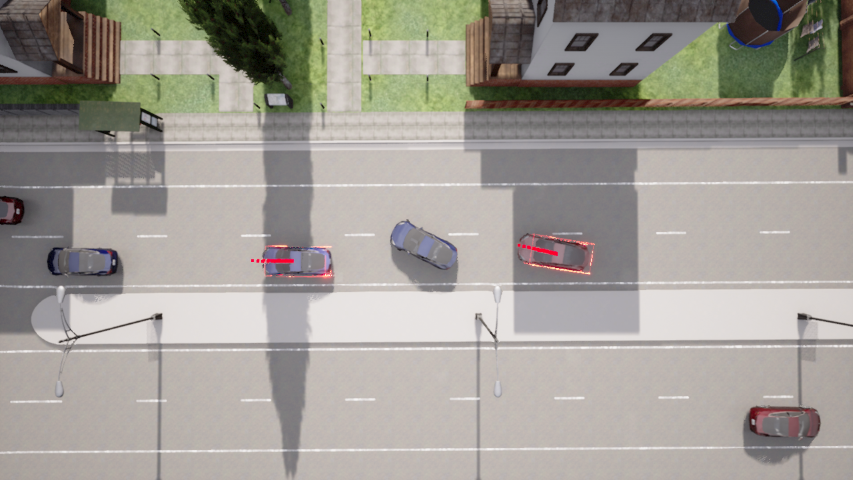}
\label{fig:Roundabout1}
}
\subfigure[Roundabout: $t=6.30$\,s]{
\includegraphics[width=0.23\linewidth]{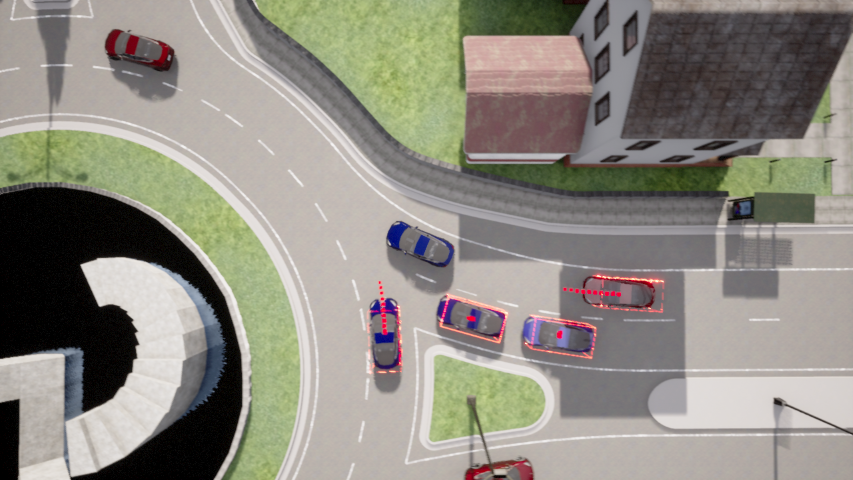}\label{fig:Roundabout2}
}
\subfigure[Roundabout: $t=9.55$\,s]{\includegraphics[width=0.23\linewidth]{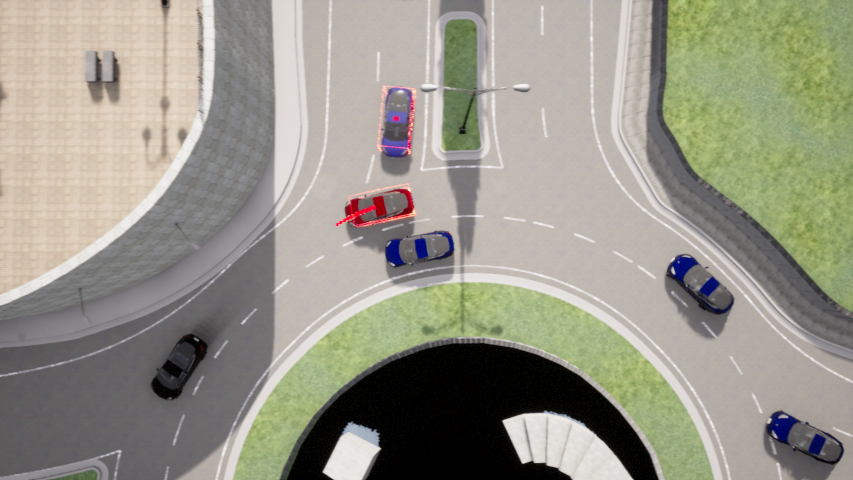}                  
\label{fig:Roundabout3}}
\subfigure[Roundabout: $t=16.15$\,s]{
\includegraphics[width=0.23\linewidth]{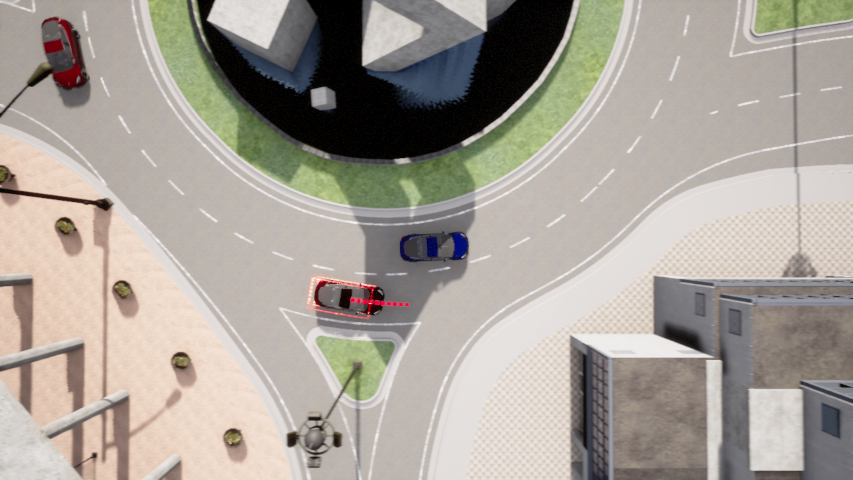}
\label{fig:Roundabout4}
}
\subfigure[Sig-Inter: $t=1.00$\,s]{
\includegraphics[width=0.23\linewidth]{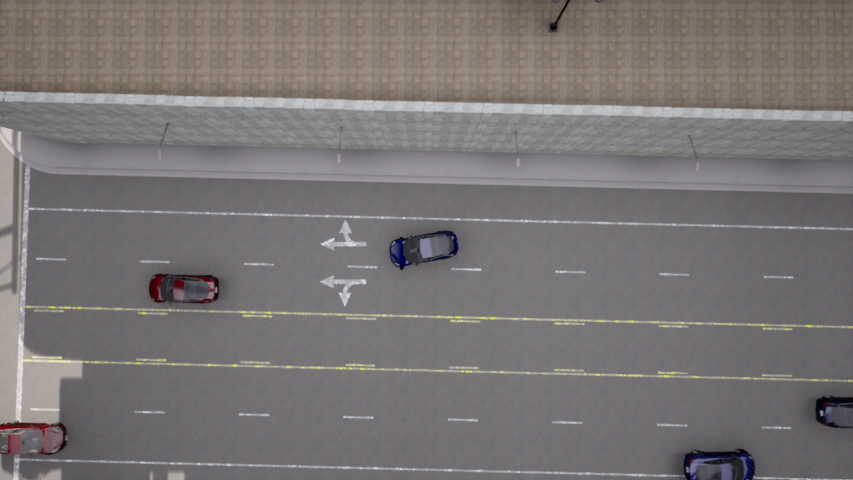}
\label{fig:sig-inter1}
}
\subfigure[Sig-Inter: $t=4.95$\,s]{
\includegraphics[width=0.23\linewidth]{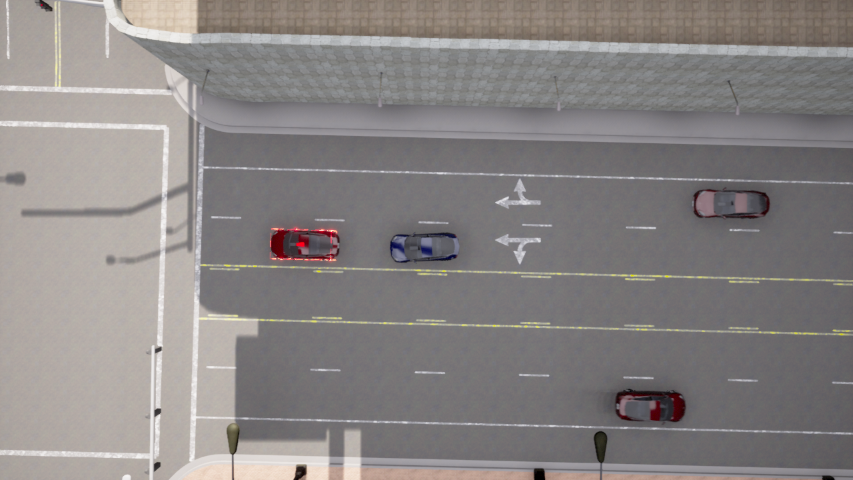}\label{fig:sig-inter2}
}
\subfigure[Sig-Inter: $t=8.65$\,s]{\includegraphics[width=0.23\linewidth]{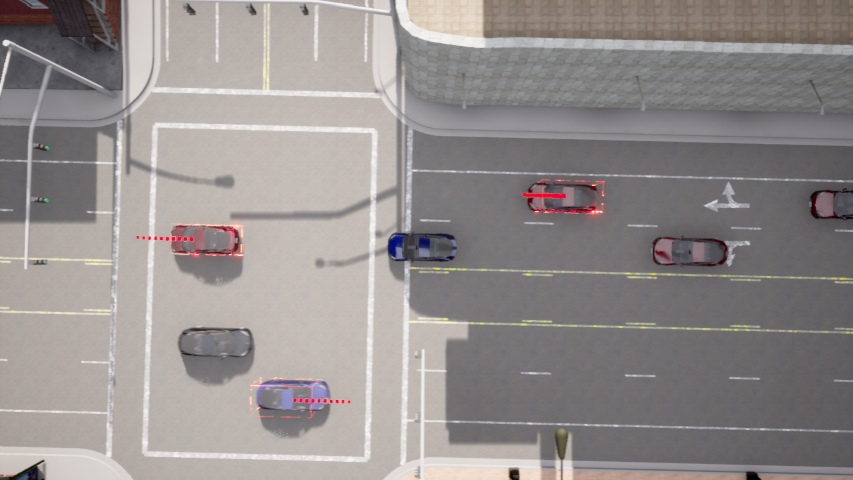}                  
\label{fig:sig-inter3}}
\subfigure[Sig-Inter: $t=16.25$\,s]{
\includegraphics[width=0.23\linewidth]{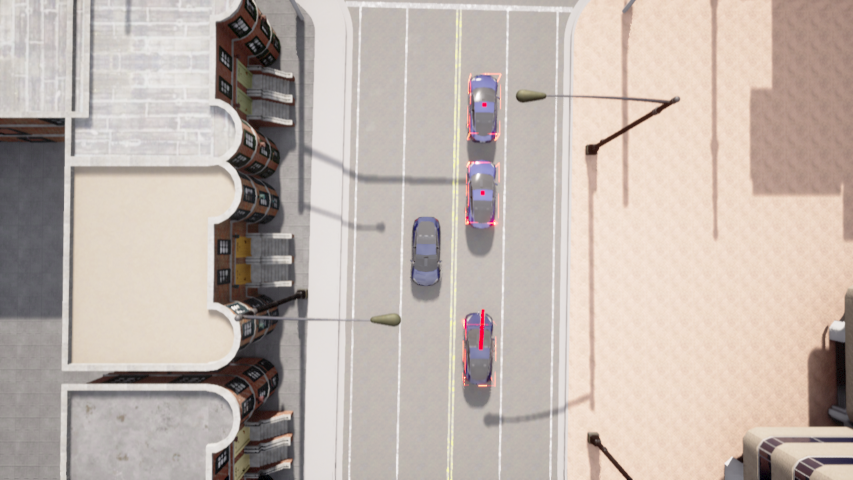}
\label{fig:sig-inter4}
}
\subfigure[Mix-T-U: $t=10.60$\,s]{
\includegraphics[width=0.23\linewidth]{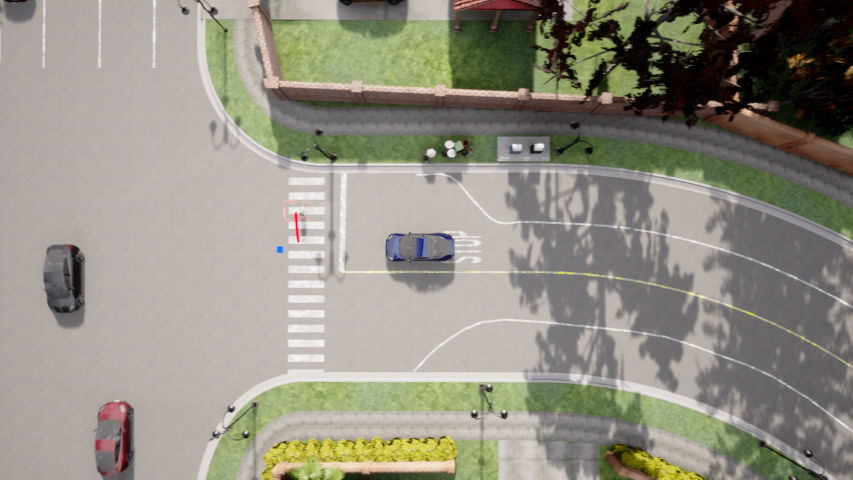}
\label{fig:mixTU1}
}
\subfigure[Mix-T-U: $t=12.05$\,s]{
\includegraphics[width=0.23\linewidth]{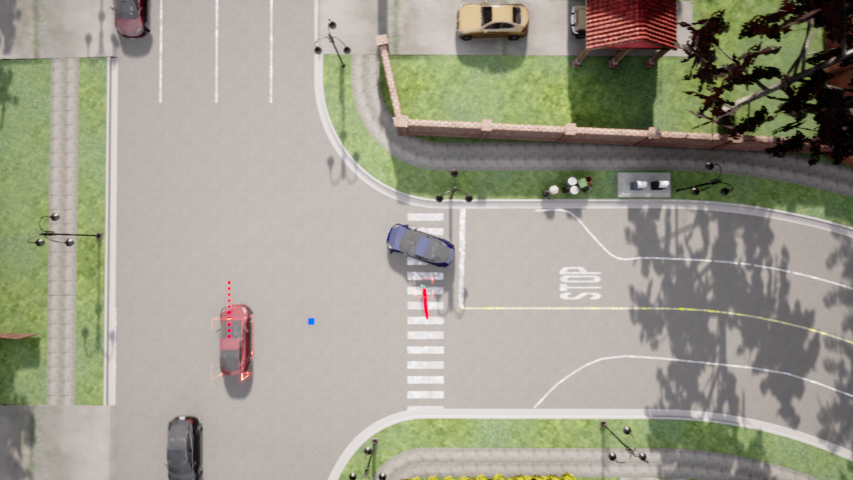}\label{fig:mixTU2}
}
\subfigure[Mix-T-U: $t=14.00$\,s]{\includegraphics[width=0.23\linewidth]{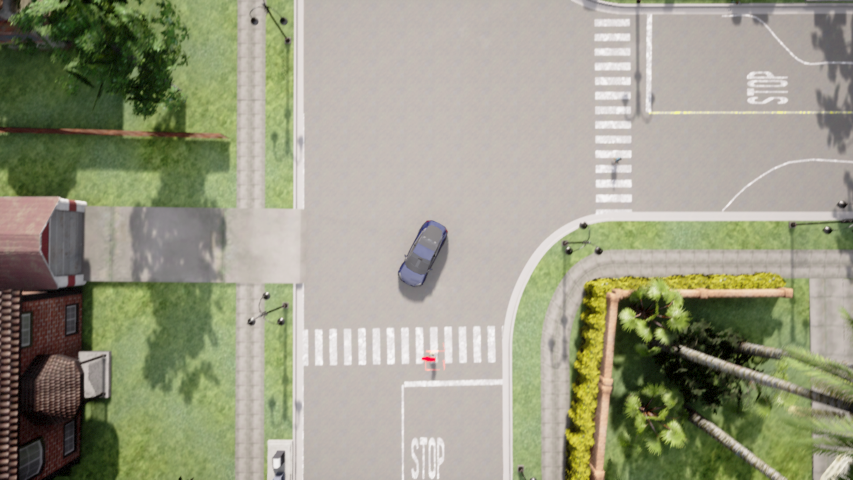}                  
\label{fig:mixTU3}}
\subfigure[Mix-T-U: $t=15.90$\,s]{
\includegraphics[width=0.23\linewidth]{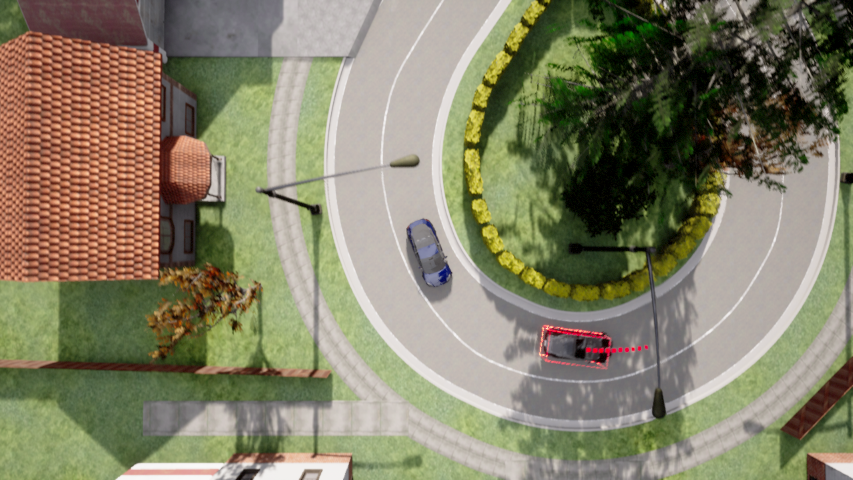}
\label{fig:mixTU4}
}
\caption{{Simulation results by UDMC in four illustrative urban driving scenarios in CARLA. Four keyframes for each scenario are selected. The red dots above the {surrounding vehicles} are the predicted states of their future trajectory in the next 10 time steps generated by IGPR. In the \textit{Mix-T-U} scenario, two pedestrians are crossing crosswalks on the way of the autonomous vehicle. Corresponding videos are accessible at \href{https://www.youtube.com/watch?v=jftTsf1jXjU}{https://www.youtube.com/watch?v=jftTsf1jXjU}.}
}
\label{fig:allStaticsSimulation}
\end{figure*}

\subsection{Environmental Settings}

We implement the proposed autonomous driving framework and algorithms in Ubuntu 20.04 LTS, and execute the simulation using AMD Ryzen 5600G CPU and NVIDIA RTX 3060 GPU with 24 GB RAM and 12 GB Graphic Memory. The OCP is solved using the Interior Point Optimizer (IPOPT) with CasADi~\cite{Andersson2019} built in Python.
In the CARLA 0.9.14 simulator, the {AV} {is controlled by throttle, brake, and steering, which are transformed by PID controller with a low-pass filter for a smoother control input from the acceleration and steering values generated by MPC.} The {surrounding vehicles} are driven by the embedded \texttt{autopilot} for a {closed-loop evaluation}. 

We design the following four challenging scenarios to verify the effectiveness of the proposed UDMC {in a targeted manner}.
\subsubsection{Scenario 1: Multi-Lane Adaptive Cruise Control (ML-ACC)}
In Fig.\ref{fig:MLACC1}-Fig.\ref{fig:MLACC4}, the AV maintains its lane at $v_\text{acc}=45$\,km/h, keeping a safe distance from nearby vehicles. It encounters four vehicles, two in each lane, requiring overtaking for efficient driving.

\subsubsection{Scenario 2: Roundabout}
Fig.\ref{fig:Roundabout1}-Fig.\ref{fig:Roundabout4} depict the AV entering and exiting a roundabout at $v_\text{ra} = 40$\,km/h. The scenario involves merging with other vehicles, navigating lane changes, and complex driving dynamics.

\subsubsection{Scenario 3: Signalized Intersection (Sig-Inter)}
In Fig.\ref{fig:sig-inter1}-Fig.\ref{fig:sig-inter4}, the AV approaches a left turn at a busy intersection with a traffic light about to change after $T_\text{tl} = 4.5$\,s at $v_\text{si} = 25$\,km/h, requiring swift responses to changing signals.

\subsubsection{Scenario 4: Mixed-Traffic T-Junction with U-Turn (Mix-T-U)}
Fig.\ref{fig:mixTU1}-Fig.\ref{fig:mixTU4} illustrate a T-junction scenario with pedestrians and vehicles interacting. The AV navigates through this dynamic environment at $v_\text{mtu}=35$\,km/h, encountering complex traffic interactions.
\subsection{Preparation for UDMC Execution}
\subsubsection{Parameter Identification}

\begin{table}[t]
  \centering
  \caption{Value of the Identified Parameters}
  \resizebox{1.0\linewidth}{!}{
    \begin{tabular}{cccccc}
    \toprule
    Notation & Value & Unit & Notation & Value & Unit \\
    \midrule
    $k_f^*$    & -102129.83 & N/rad & $l_r^*$    & 1.603  & m\\
     $k_r^*$    & -89999.98 & N/rad &$m^*$     & 1699.98  & kg \\
    $l_f^*$    & 1.287  & m& $I_z^*$    & 2699.98 & $\text{kg}\cdot \text{m}^2$\\
    \bottomrule
    \end{tabular}%
    }
  \label{tab:param_model}%
\end{table}%
{The vehicle dynamics model explains how a vehicle reacts to inputs like acceleration and steering. By determining the model parameters, an accurate representation of vehicle behavior is ensured. This is vital for developing control inputs that effectively predict and manage the motion of the {AV}.} To identify the parameters of the vehicle dynamics model, we perform an optimization process using the SLSQP algorithm. According to the vehicle dynamics model (\ref{eq:bicycle_dynamics}), the identified vehicle model can be expressed as:

\begin{equation}
    \hat{\bm {x}}_{\tau+1}=f(\bm x_\tau, \bm u_\tau, \bm p_e),
\end{equation}
where $\bm p_e = [k_f, k_r, l_f, l_r, m, Iz]^T$ denotes the parameter vector to be estimated.
The optimization process aims to minimize the residual function $R(\bm p_e)$ of the vehicle state:
\begin{equation}
\label{eq:param_est}
\begin{array}{ll}
\underset{{\boldsymbol{p_e}}}{\min} & R(\bm p_e) = \sum_{i = 1}^{M - 1} |\bm x_{i + 1} - \hat{\bm {x}}_{\tau+1}| \\
\text { s.t.} & l_f + l_r = 2.89 \\
& l_f, l_r \geq 0\\
& 1000 \leq m \leq 2200 \\
& I_z \geq 0\\
& k_f, k_r \geq 0,
\end{array}
\end{equation}
where $M$ denotes the total number of data points. The data for parameter identification is gathered during the {AV} autonomous driving using \texttt{autopilot} mode. To ensure that the estimated parameters are within reasonable bounds, we imposed several constraints on the optimization process as shown in (\ref{eq:param_est}).

The optimization process results in the identified parameter values in Table~\ref{tab:param_model}. With these estimated parameters, the vehicle dynamics model can be used as the dynamics constraint for {AV}'s decision-making and control in the OCP\,(\ref{NMPCOptProb}). 

\subsubsection{{Motion Prediction of Traffic Participants}}
To fully take advantage of the proposed UDMC scheme, we need to provide not only accurate but also efficient motion prediction {of {surrounding vehicles} and pedestrians to formulate their PFs}. This section leverages the proposed IGPR in Section~\ref{subsec:SVPredictIGPR}. {Considering the motion prediction of {surrounding vehicles}}, first, we collect 400 pieces of historical state vectors $\bm p^i_\tau = [x^i_\tau, y^i_\tau, \varphi^i_\tau, v^i_{x,\tau}, v^i_{y,\tau}]^T\in\,\mathbb R^5 (i \in \{j|j\in \mathbb N, j<100\})$ of 100 vehicles driving by \texttt{autopilot} embedded in CARLA. Second, we pre-process the original driving data to be centered from the origin of the map frame and flatten the state matrices $S_i\in \mathbb R^{25\times 5}$ to be vectors $\bm z_i \in \mathbb R^{75}$ and $\bm y_i \in \mathbb R^{30}$, where $i \in \{0,1,\ldots,m\}$. Notice that we drop the last two elements $[v_x, v_y]$ of the predicted/output states from the collected data pieces to make the IGPR model more concise and efficient. Third, the Gram matrices $\bm {\Phi_{ZZ}}^i$ are computed by (\ref{eq:gramMatCal}) along with maximizing the likelihood with 10 restarts for the optimizer to find the optimal hyperparameters. After the above training process, the IGPR is ready to predict the surrounding vehicles' future states $\bm p_\tau, \forall \tau \in \{1,2,\ldots, 10\}$. {The same procedure applies to the prediction of pedestrian movement.}
\begin{table}[t]
  \centering
  \caption{Parameters of the Designed APF}
  \resizebox{1.0\linewidth}{!}{
    \begin{tabular}{cccccc}
    \toprule
    Parameter & Value & Parameter & Value & Parameter & Value \\
    \midrule
    $a_\text{NR}$ & 100.0   & $a_\text V$ & 500.0   & $w_R$  & 3.5 \\
    $b_\text{NR}$ & 2.0     & $b_\text V$ & 1.0     & $a_\text{TL1}$ & 200.0 \\
    $a_\text{CR}$ & 10.0    & $r_a$ & 2.4   & $a_\text{TL2}$ & 1000.0 \\
    $b_\text{CR}$ & 0.5   & $r_b$ & 1.0     & $r_\text{offset}$ & 0.25 \\
    \bottomrule
    \end{tabular}%
    }
  \label{tab:param}%
\end{table}%

\begin{table*}[htbp]
  \centering
  \caption{{Comparison of Driving Performance in Urban Driving Scenarios}}
    \begin{tabular}{cccccccc}
    \toprule
    Method & Scenario & Comp. Time (ms) & Col. & TRV & IB & Dur. for TTC $<$ 1.5 (s) & Travel Time (s)\\
    \midrule
    \multirow{4}[2]{*}{FSM with PID~\cite{dosovitskiy2017carla} (rule-based)} & ML-ACC & $10.61\pm3.57$ & 0     & 0     & 0     & 0.10 (13.5\%) & 26.6 \\
          & Roundabout & $8.93\pm2.85$ & 2     & 3     & 2     & 0.55 (15.5\%) & 35.6 \\
          & Sig-Inter & $12.89\pm3.76$ & 1     & 0     & 1     & 0.65 (47.7\%) & 18 \\
          & Mix-Traf & $5.14\pm0.64$ & 1     & 0     & 0     & \textbackslash{} & \textbackslash{} \\
    \midrule
    \multirow{4}[2]{*}{InterFuser~\cite{shao2023safety} (learning-based)} & ML-ACC &  $178.20\pm 23.78$    &  0     &  0     & 0     &  0 (10.73\%)       &   52.9 \\  
          & Roundabout &  $179.98\pm 22.62$    &  0     &   0    &    0   & 0 (6.5\%)      &  52.85 \\ 
          & Sig-Inter &     $180.33\pm 25.28$     &   0    &   0    &   0       & 0 (16.8\%)       &  28.95\\
          & Mix-T-U &  $162.74\pm 30.41$     &    0   &   0    &   0   &   0 (0\%)      & 45.75 \\
    \midrule
    \multirow{4}[2]{*}{RSS with PID~\cite{shalev2017formal} (reactive-based)} & ML-ACC &   $11.62\pm3.14$    & 0     & 0     & 0     & 0.15 (13.3\%) & 26.7 \\
          & Roundabout &  $9.67\pm3.22$  & 1     & 0     & 0     & 0.85 (9.93\%) & 37.75 \\
          & Sig-Inter &   $13.29\pm3.48$    & 0     & 0     & 0     & 0.20 (27.8\%) & 20.5 \\
          & Mix-Traf &   $6.49\pm 9.81$    & 1     & 2     & 0     & 0 (1.0\%) & 31.25 \\
    \midrule
    \multirow{4}[2]{*}{UDMC (ours)} & ML-ACC & $35.07\pm13.00$ & 0     & 0     & 0     & 0 (23.12\%) & 16.65 \\
    & Roundabout & $30.35\pm12.68$ & 0     & 0     & 0     & 0 (7.02\%) & 17.8 \\
    & Sig-Inter & $25.44\pm14.32$ & 0     & 0     & 0     & 0 (20.30\%) & 16.5 \\
    & Mix-Traf & $27.32\pm11.34$ & 0     & 0     & 0     & 0 (37.9\%) & 24.75 \\
    \bottomrule
    \end{tabular}%
  \label{tab:CompScenariosMethods}%
\end{table*}%

\subsection{Comparative Study}\label{subsec:drivePerformComp}
We design the following two sets of comparisons to evaluate the proposed UDMC scheme's performance and verify the effectiveness of the environment description by APF and the motion prediction of {traffic participants} by IGPR.

First, we compare UDMC with 1) a rule-based hierarchical driving algorithm, FSM with PID~\cite{dosovitskiy2017carla}, 2) a learning-based state-of-the-art end-to-end integrated driving model, InterFuser (ranks No.\,1 on the CARLA Leaderboard until July 2023)~\cite{shao2023safety}, {and a reactive-based driving algorithm, RSS with PID~\cite{shalev2017formal}}.

We follow the pipeline demonstrated in Section~\ref{sec:pipeline} to implement the proposed UDMC framework. For the balance of computing efficiency and driving optimality, the prediction horizon in (\ref{NMPCOptProb}) is set as $N = 10$. As the parameters of the dynamics model are identified with a time interval of $0.05$\,s, the sampling time in (\ref{eq:bicycle_dynamics}) is also set as $T_s = 0.05$\,s. The parameters of the PFs defined in Section~\ref{subsec:4-1APFDesign} are listed in Table~\ref{tab:param}. Note that the above parameters are well-fitted to all kinds of urban driving scenarios because the size of the vehicles and the width of road lanes has universal standards and do not change much in the transportation systems in different regions. Hence, the above parameters are versatile and universal.

In each scenario, we deploy the aforementioned methods, and the testing results are demonstrated in Table~\ref{tab:CompScenariosMethods}, which presents a comprehensive comparison of driving performance for the different methods.
{In terms of the evaluation matrices in Table~\ref{tab:CompScenariosMethods},} \textit{Comp. Time} denotes the computational time of executing one step of the AV. If the {AV} collides with a surrounding vehicle, the collision index (\textit{Col.}) shall rise by one. {Traffic rule violation (\textit{TRV}) means that the {AV} drives out of the road, crosses non-crossable lane markings, or runs a red traffic light.} {Impolite behavior (\textit{IB}) means the {surrounding vehicles} are ``hindered'' by the {AV}, although there is no collision between the {AV} and {surrounding vehicles}.} {In our simulation, all the {surrounding vehicles} are operating in the built-in \texttt{autopilot} mode. Consequently, we consider any sudden braking of an {surrounding vehicle} resulting from the motion of an {AV} as an indication of the {AV}'s impolite behavior. For instance, if an {AV} changes lanes without maintaining a safe distance from the vehicle behind it, the following vehicle has to decelerate or abruptly brake in order to avoid a collision. In such cases, the \texttt{autopilot} system always opts for sudden braking as the preferred response to this situation. For TTC, we set the thresholds as 1.5\,s~\cite{yan2024evaluation}. If the metric is larger than this, it is considered an alarm. Note that the percentage listed after each time duration of TTC in Table~\ref{tab:CompScenariosMethods} {and Table~\ref{tab:ablation}} represents the proportion of valid data recording periods in relation to the entire driving process.} {Valid data means the data recorded with a valid leading vehicle of the AV, which reflects the environmental changes during closed-loop testing.}

As depicted in Table~\ref{tab:CompScenariosMethods}, the first method, FSM with PID, shows relatively low computation time but has a few instances of collisions, traffic rule violations, and impolite behavior across the four scenarios. {Moreover, this method fails to guide the {AV} to the destination in the last scenario because of the deadlock with a pedestrian.} Besides, InterFuser records no collisions or traffic rule violations but exhibits considerably higher computation time and shows a longer travel time in most scenarios than other methods. {Moreover, the TTC {(including the percentage value)} for InterFuser also implies this method is too conservative. Thus, this method does not achieve a balance of safety and efficiency.} {Moreover, RSS with PID method performs safer than FSM, while keeping a competent driving efficiency to FSM. However, the TTC of this method is significant, which means that the safety margin is small and prone to collide with the leading vehicle. Lastly, it is significant that our proposed UDMC scheme achieves the best traveling efficiency with the largest safety margin with the leading vehicle. Furthermore, the driving performance of our method is best as well, without any collision, traffic rule violation, or impolite behaviors, while maintaining real-time computational efficiency.}

\begin{figure}[t]
\centering
\subfigure[PF value in Scenario 1]{
\includegraphics[width=0.45\linewidth]{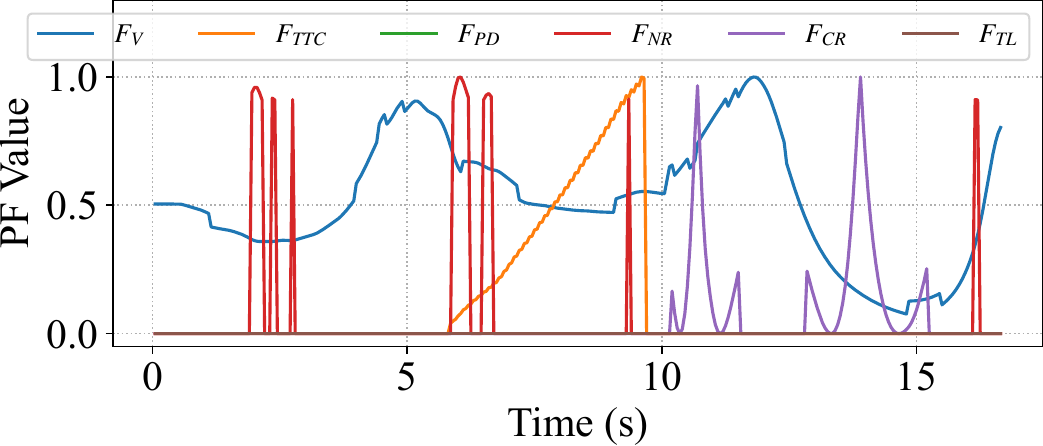}
\label{fig:apf_s1}
}
\subfigure[PF value in Scenario 2]{
\includegraphics[width=0.45\linewidth]{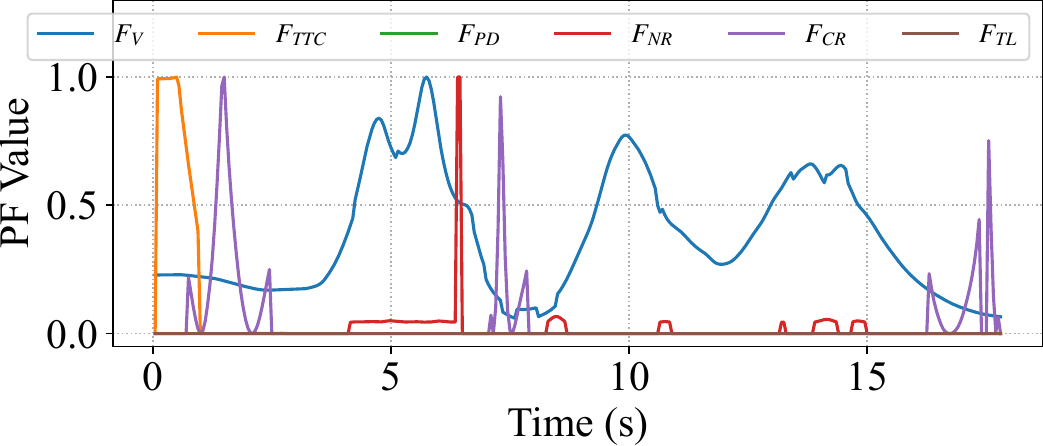}
\label{fig:apf_s2}
}
\subfigure[PF value in Scenario 3]{
\includegraphics[width=0.45\linewidth]{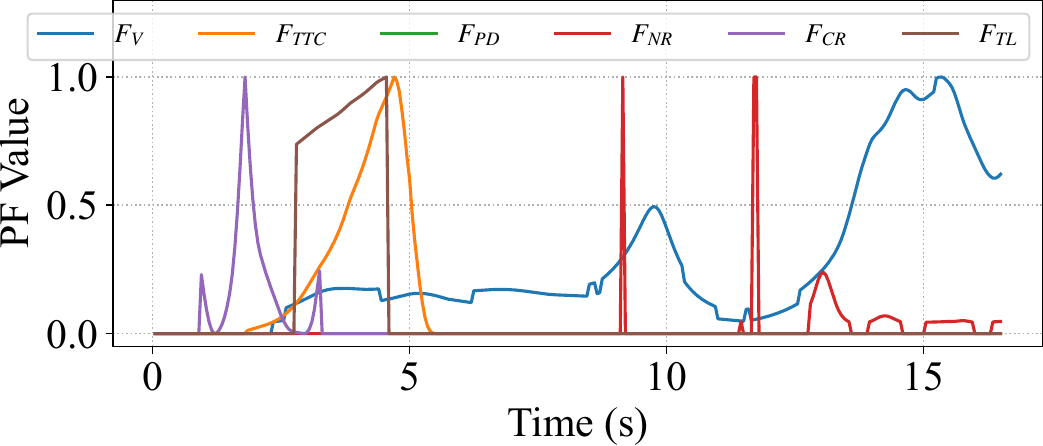}
\label{fig:apf_s3}
}
\subfigure[PF value in Scenario 4]{
\includegraphics[width=0.45\linewidth]{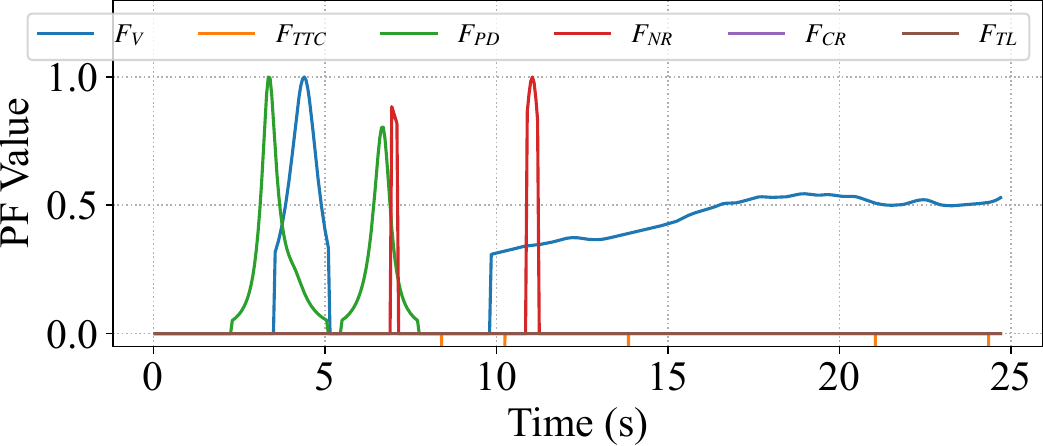}
\label{fig:apf_s4}
}
\caption{PF value in the four urban driving scenarios. The timeline of the sub-figures corresponds to the video frames in the simulation results of Fig.~\ref{fig:allStaticsSimulation}.}
\label{fig:APF_value}
\end{figure}
\begin{figure}[t]
\centering
\subfigure[Control input in Scenario 1]{
\includegraphics[width=0.46\linewidth]{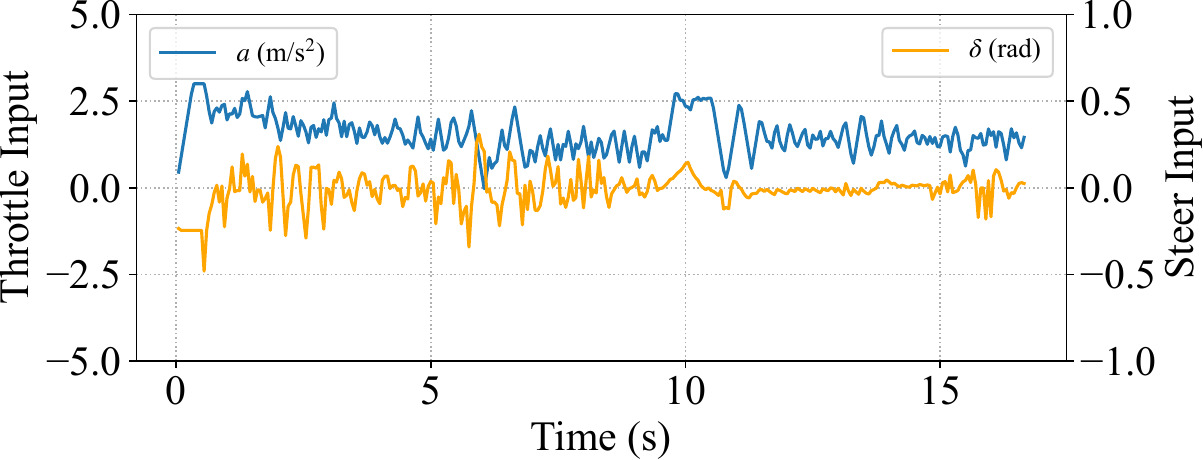}
\label{fig:ctrlInput_s1}
}
\subfigure[Control input in Scenario 2]{
\includegraphics[width=0.46\linewidth]{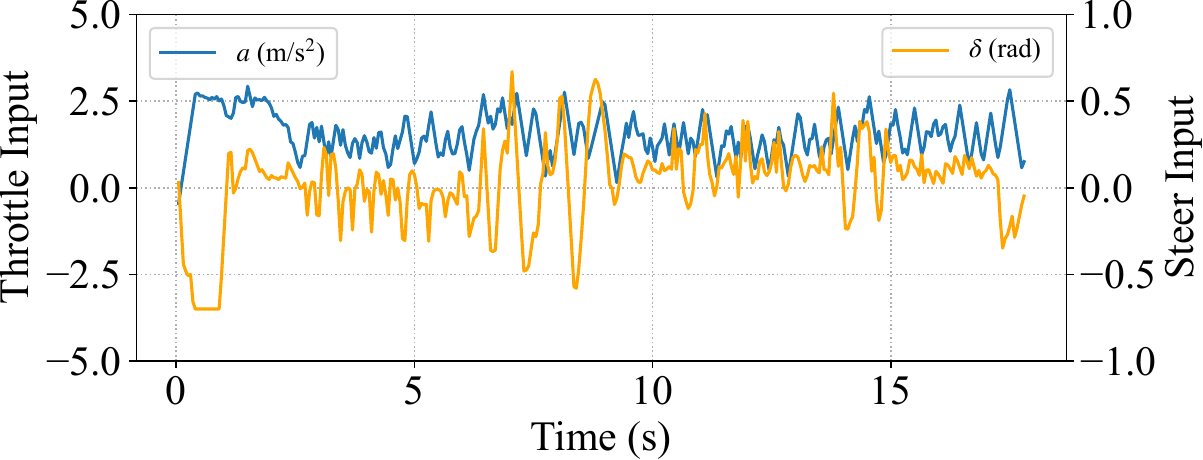}
\label{fig:ctrlInput_s2}
}
\subfigure[Control input in Scenario 3]{
\includegraphics[width=0.46\linewidth]{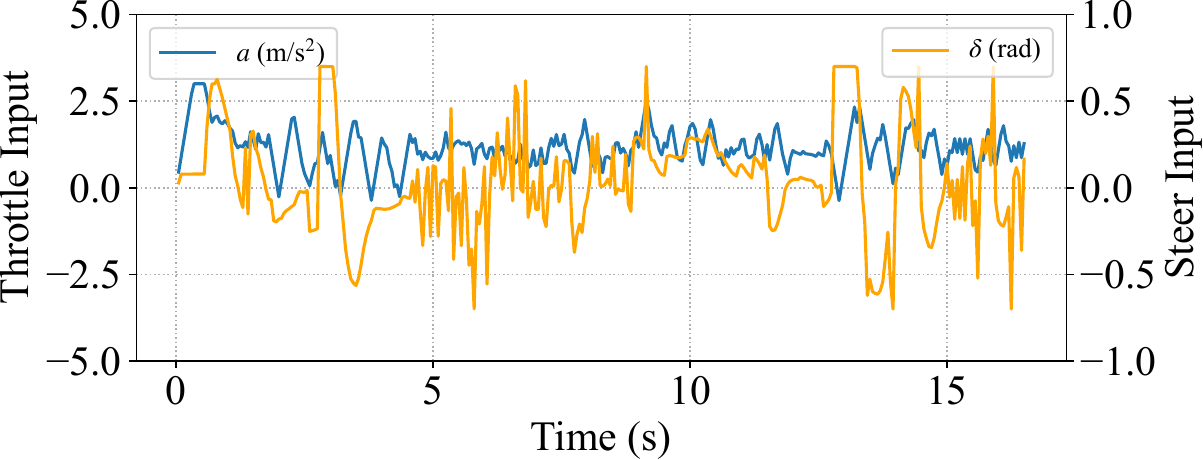}
\label{fig:ctrlInput_s3}
}
\subfigure[Control input in Scenario 4]{
\includegraphics[width=0.46\linewidth]{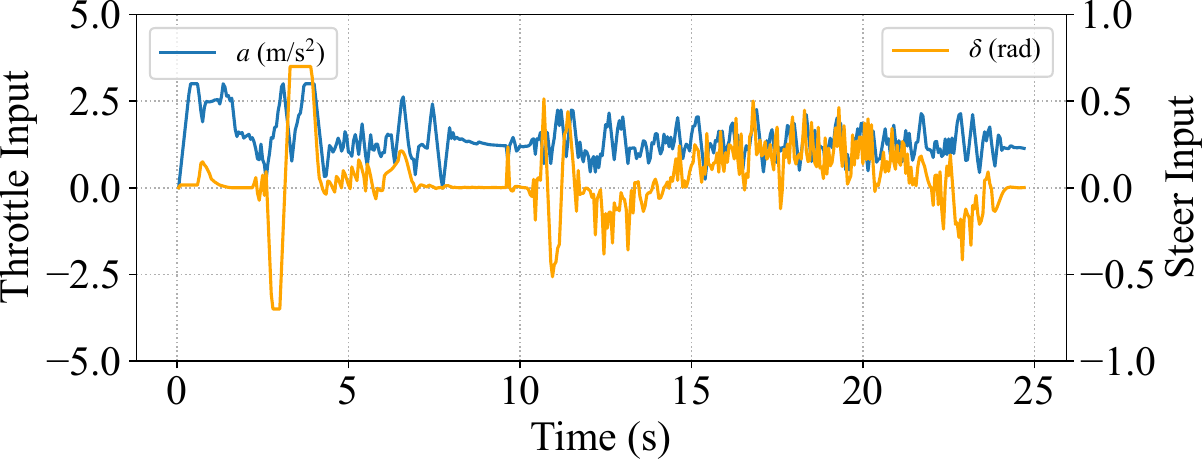}
\label{fig:ctrlInput_s4}
}

\caption{{Control input in the four urban driving scenarios. The left and right vertical axes represent the acceleration and steering angle, respectively.}}
\label{fig:controlInput}
\end{figure}

{The reasons for the distinguished performance of our UDMC framework are analyzed as follows. The unified OCP comprises a variety of PFs to model the traffic objects for the AV's decision and control.}
To illustrate the utility of the proposed PFs, we select four keyframes for visualization in Fig.~\ref{fig:allStaticsSimulation} from the simulations in Section~\ref{subsec:drivePerformComp} along with their PF values in Fig.~\ref{fig:APF_value}.
One can understand the utility of the PFs against the {keyframes} and the corresponding PF values. For Scenario 1, as shown in Fig. \ref{fig:MLACC1}, the leading vehicle maintains a lower speed, so that the {AV} slows down 
and changes into its left neighbor lane temporarily at around $10$\,s to maintain the desired velocity. And when no {surrounding vehicle} is occupying the right lane, the {AV} returns to it at about $14$\,s to suppress the tracking error. During the aforementioned process, {as depicted by Fig.~\ref{fig:apf_s1},} two peaks of $F_\text{V}$ are caused by the lateral surrounding vehicles' occurrence{, while one peak of $F_\text{TTC}$ is detected to urge the AV to keep a safe margin from the leading vehicle.} Due to the mighty PF of non-crossable lane markings $F_\text{NR}$ {around 12\,s}, the {AV} avoids driving out of the road {complying with the traffic rules and ensuring driving safety}. Besides, $F_\text{CR}$ raises twice at around $10$\,s and $15$\,s, which indicates the lane-changing behavior of the {AV}.

In Scenario 2, the {AV} aims to get through the roundabout with interaction with other vehicles. {As shown in Fig.~\ref{fig:Roundabout1},} after a lane change behavior around 1\,s, the {AV} enters the roundabout at 6\,s, while there are {surrounding vehicles} on the left side. 
{From Fig.~\ref{fig:apf_s2}, the three peaks of $F_\text{CR}$ imply that the {AV} made lane change maneuver three times, under the continuous influence from the {surrounding vehicles} $F_\text{V}$.}
{During the roundabout driving process, the {AV} drives calmly in the roundabout until driving out of it, without being interfered with by the merging-in and driving-out {surrounding vehicles}.}

\begin{remark}
    {For FSM and InterFuser, the {AV} stops at the stop line to wait for the {surrounding vehicle} to pass first. But as demonstrated, the {AV} with UDMC keeps its velocity and enters the roundabout without hindering the left vehicle. This decision exhibits the flexibility and versatility of UDMC.}
\end{remark}

For Scenario 3, as shown in {Fig. \ref{fig:sig-inter2}}, when the traffic light is red at the beginning, {the {AV} first follows its guidance waypoints {to drive to the far left lane}, and smoothly stops as a result of the combination of the lateral and longitudinal forces from $F_\text{TL}$ and $F_\text{V}$, as shown in Fig.~\ref{fig:apf_s3}.} {Notice that due to the proposed VPF, although the right lane is empty and the adjacent right lane marking is crossable, the {AV} stays in the left-turning lane with traffic rule compliance.} After the traffic light turns green at $4.5$\,s, the {AV} gradually increases velocity to cross the intersection and drives into the correct road.

{As illustrated in Fig.~\ref{fig:mixTU1}, in Scenario 4, the {AV} begins at the T-junction with pedestrians walking on the two crosswalks. At around 3\,s and 7\,s, the value of PF for pedestrian $F_\text{PD}$ climbs distinctly as shown in Fig.~\ref{fig:apf_s4} to urge the {AV} to make avoidance actions. Thereafter, the {AV}, behind a leading vehicle, enters a U-turn road section with large curvature. The PF values of $F_\text{V}$ after 10\,s in Fig.~\ref{fig:apf_s4} keep a medium range to guide the {AV} driving safely with an appropriate following distance.} The video accompanying Figure~\ref{fig:allStaticsSimulation} illustrates that, as depicted in Figure~\ref{fig:controlInput}, the autonomous vehicle's control exhibits predominantly smooth behavior in the majority of instances.

\begin{table}[t]
  \centering
  \caption{{Closed-loop evaluation on CARLA Town05 Short Benchmark.}
}
    \begin{tabular}{cccc}
    \toprule
    \multirow{2}[2]{*}{Method} & Driving & Route & Infraction \\
          & Score $\uparrow$ & Completion $\uparrow$ & Rate $\uparrow$ \\
    \midrule
    Transfuser~\cite{prakash2021multi} & 54.5 & 78.4 & 0.70\\
    LBC~\cite{chen2020learning} & 31.0 & 55.0 & 0.56 \\
    ST-P3~\cite{hu2022st} & 55.1  & 86.7  & 0.64 \\
    VAD~\cite{jiang2023vad}   & 64.3  & 87.3  & 0.74 \\
    UDMC (ours)  & 76.5  & 85.7  & 0.89 \\
    \bottomrule
    \end{tabular}%
  \label{tab:carlaTown05Benchmark}%
\end{table}%

{To further evaluate the effectiveness of our method, as shown in Table~\ref{tab:carlaTown05Benchmark}, we conducted experiments using the CARLA Town05 Short Benchmark, which encompasses a diverse range of driving scenarios such as varying traffic densities, different road layouts, and a variety of vehicle types and sizes. For a fair comparison, we adapt our algorithm from CARLA version 0.9.14 to version 0.9.10. It is important to note that the performance results obtained in this evaluation may not fully reflect the true capabilities of the UDMC, as our approach prioritizes the shortest path during global planning, sometimes resulting in being flagged as ``out of route" (detailed analysis provided in Section~\ref{sec:hazardAnalysis}). {Despite the limitations of the simulator, our method consistently outperforms the four representative algorithms, especially in terms of driving score and infraction rate.} It is worth mentioning that while other methods rely on sensor-based approaches, our UDMC currently leverages V2X communication. Despite this difference, all {five} methods share a common goal of facilitating urban autonomous driving without specific scenario preferences.}

\begin{table*}[t]
  \centering
  \caption{{Ablation Study of the UDMC Framework}}
  \resizebox{1.0\linewidth}{!}{
    \begin{tabular}{ccccccccccc}
    \toprule
    Ablation & Prediction & SV model & Safe. Enh. & Scenario & Comp. Time (ms) & Col. & TRV & IB & Dur. for TTC $<$ 1.5 (s) & Travel Time (s) \\
    \midrule
    \multirow{4}[2]{*}{UDMC w/o Pred.} & \multirow{4}[2]{*}{N/A} & \multirow{4}[2]{*}{Ellipse} & \multirow{4}[2]{*}{N/A} & ML-ACC & $25.30\pm10.58$ & 0     & 0     & 2     & 0.25 (9.6\%) & 16.5 \\
          &       &       &       & Roundabout & $21.09\pm10.09$ & 1     & 0     & 1     & 0 (3.82\%) & 22.25 \\
          &       &       &       & Sig-Inter & $21.89\pm11.59$ & 0     & 0     & 0     & 1.55 (19.26\%) & 16.35 \\
          &       &       &       & Mix-Traf & $13.88\pm5.78$ & 1     & 0     & 0     & \textbackslash{} & \textbackslash{} \\
    \midrule
    \multirow{4}[2]{*}{UDMC with LSTM} & \multirow{4}[2]{*}{LSTM} & \multirow{4}[2]{*}{Ellipse} & \multirow{4}[2]{*}{N/A} & ML-ACC & $131.72\pm35.54$ & 0     & 0     & 0     & 0.55 (26.36\%) & 17.45 \\
          &       &       &       & Roundabout & $85.10\pm31.49$ & 0     & 1     & 1     & 0 (4.4\%) & 18.3 \\
          &       &       &       & Sig-Inter & $139.49\pm90.83$ & 0     & 1     & 0     & 1.45 (34.2\%) & 16.05 \\
          &       &       &       & Mix-Traf & $53.49\pm39.10$ & 1     & 0     & 1     & 0 (34.6\%) & 25 \\
    \midrule
    \multirow{4}[2]{*}{UDMC with DC} & \multirow{4}[2]{*}{IGPR} & \multirow{4}[2]{*}{Dou. Circ.} & \multirow{4}[2]{*}{N/A} & ML-ACC & $35.38\pm14.80$ & 0     & 0     & 1     & 0.95 (16.6\%) & 16.8 \\
          &       &       &       & Roundabout & $31.10\pm11.11$ & 0     & 0     & 1     & 0 (5.26\%) & 18.05 \\
          &       &       &       & Sig-Inter & $27.03\pm13.15$ & 0     & 1     & 1     & 1.75 (82.4\%]) & 16 \\
          &       &       &       & Mix-Traf & 22.08+9.18 & 0     & 0     & 1     & 0.05 (20.19\%) & 25.25 \\
    \midrule
    \multirow{4}[2]{*}{UDMC w/o TTC} & \multirow{4}[2]{*}{IGPR} & \multirow{4}[2]{*}{Ellipse} & \multirow{4}[2]{*}{N/A} & ML-ACC & $36.06\pm15.02$ & 0     & 0     & 0     & 0.75 (24.4\%) & 16.75 \\
          &       &       &       & Roundabout & 30.01+11.83 & 0     & 0     & 0     & 0 (4.5\%) & 17.7 \\
          &       &       &       & Sig-Inter & $29.20\pm13.65$ & 0     & 0     & 0     & 1.45 (21.6\%) & 15.5 \\
          &       &       &       & Mix-Traf & $27.32\pm11.34$ & 0     & 0     & 1     & 0 (38.7\%) & 25.05 \\
    \midrule
    \multirow{4}[2]{*}{UDMC (full version)} & \multirow{4}[2]{*}{IGPR} & \multirow{4}[2]{*}{Ellipse} & \multirow{4}[2]{*}{$F_\text{TTC}$} & ML-ACC & $35.07\pm13.00$ & 0     & 0     & 0     & 0 (23.12\%) & 16.65 \\
          &       &       &       & Roundabout & $30.35\pm12.68$ & 0     & 0     & 0     & 0 (7.02\%) & 17.8 \\
          &       &       &       & Sig-Inter & $25.44\pm14.32$ & 0     & 0     & 0     & 0 (20.30\%) & 16.5 \\
          &       &       &       & Mix-Traf & $27.32\pm11.34$ & 0     & 0     & 0     & 0 (37.9\%) & 24.75 \\
    \bottomrule
    \end{tabular}%
    }
  \label{tab:ablation}%
\end{table*}%

\subsection{Ablation Study}
{As shown in Table~\ref{tab:ablation}, we executed the ablation simulations based on three factors, prediction module, modeling for surrounding vehicles (\textit{SV model}), and the existence of safety enhancement (\textit{Safe. Enh.}) to verify the necessity of all the parts of the scheme. The detailed ablation plan is as follows.} 1) \textit{UDMC w/o Pred.}: the prediction of {traffic participants} is disabled and current states of {surrounding vehicles} are directly sent to the OCP. 2) \textit{UDMC with LSTM}: LSTM is used to predict the surrounding vehicles' future positions in 10 time steps with 50\,ms time interval. 3) \textit{UDMC with DC}: the ellipsoid is replaced with double circles in $F_\text{V}$ to construct the PF of {surrounding vehicles}. {4) \textit{UDMC w/o TTC}: All the techniques used by UDMC are included in this version of the method, except that no $F_\text{TTC}$ is applied.}
Notice that the LSTM network uses the same training dataset as a baseline for our proposed motion prediction method based on IGPR. For the learning process of LSTM, we divide the dataset into training and testing sets by a proportion of 9:1. The LSTM network is structured with two LSTM layers consisting of 128 units each, followed by a dense layer with 50 units and ReLU activation, and an output dense layer with 30 units (corresponding to the 10 predicted waypoints $\bm p_\tau$).

The motion prediction module is necessary. 
Referring to Table~\ref{tab:ablation}, if there is no surrounding vehicles' prediction, the {AV} is prone to exhibit more impolite decisions (e.g., overtaking with a minor safety distance), which is unsafe for urban driving. {If the prediction module is deployed, a}s illustrated in Fig.~\ref{fig:allStaticsSimulation}, the {{surrounding vehicles} and pedestrians'} motion prediction results are plotted above the vehicles with red dots, which are incorporated into the OCP to construct the traffic participants' future states with the same time interval $T_s$. For instance, in Scenario 1, if there is no surrounding vehicles' prediction {(as \textit{UDMC w/o Pred.} in Table~\ref{tab:ablation})}, the {AV} drives drastically and often behaves impolitely toward the target lane with a following vehicle. In Scenario 2, the {AV} keeps a smooth driving in the roundabout as a result of accurate motion prediction of the {surrounding vehicles} in the adjacent lanes. In the intersection of Scenario 3 and Scenario 4, due to precise prediction, the {AV} can slow down before the {surrounding vehicle or pedestrian} driving in the {AV}'s lateral direction comes too close, which also improves the driving safety and efficiency compared with the third and fourth row of Table~\ref{tab:CompScenariosMethods}. However, if LSTM is applied to predict surrounding vehicles' motion, although the driving performance is relatively good, the real-time performance is compromised. {Besides, although the double-circle description of vehicles is straightforward, it is prone to break traffic rules and behave impolitely due to the depression region elaborated in Remark~\ref{remark:doubleCircles}. From the last two ablations of Table~\ref{tab:ablation}, in most cases, \textit{UDMC with DC} has higher values of TTC, which indicates a more aggressive driving style in making unsafe behaviors.} {Lastly, the PF for the TTC indicator significantly improves the safety margin of the AV and no TTC alarm is detected.}
In summary, these configurations have significant defects compared with the full version of UDMC, which proves the necessity of the designed APF (including $F_\text{TTC}$) and presented IGPR, and further verifies the robust and real-time characteristics of our UDMC framework.
{Therefore, the full version of UDMC demonstrates the best overall performance, with no collisions or traffic rule violations, low computation time, and less travel time in most cases.}

\subsection{Robustness Evaluation of the UDMC framework}

To further demonstrate the robustness of our proposed UDMC scheme for urban self-driving, we execute the simulation of 40 trials for each scenario and quantitatively evaluate the time consumption to the destination and success rate. Notice that this driving condition is more challenging than the simulation in Section~\ref{subsec:drivePerformComp}, because all the {surrounding vehicles} are generated randomly near the {AV}. In each trial, 100 {surrounding vehicles} are spawned with a minimum and maximum distance from the {AV}'s spawn point of 5\,m and 300\,m, {hence containing varying traffic densities}. {We define a successful trial as one where there are no collisions with other traffic participants and all traffic rules are complied with. Hence, the evaluation metric of \textit{success rate} is the ratio of the number of successful trials and the total number of attempts.} As shown in Table~\ref{tab:robustTest}, the success rates in all four randomly generated scenarios are higher than 90\%. Therefore, our proposed UDMC can be applied to highly dynamic scenarios with satisfactory robustness and driving performance. 
We have achieved a remarkable success rate of 97.5\% specifically in the signalized intersection scenario, with only one instance of failure occurring when two {surrounding vehicles} approached the {AV} from opposite sides, leaving an extremely limited safety margin. The two failure cases in the roundabout scenario lay in the first quarter of the roundabout by running out of the inner road boundary (where no non-crossable lane marking signal is {detected}). In the multi-lane ACC scenario, when the {AV} is approaching the T-shape section (as shown in Fig.~\ref{fig:MLACC4}), one {surrounding vehicle} suddenly breaks behind the stop line, which leaves no sufficient time for the {AV} to break. For unsignalized intersection driving, one failure case is similar to the case in Scenario 3. {The three failure cases in Scenario 4 occur by running on the solid road lines because of the uncommon merging lane marking, as shown in Fig.~\ref{fig:mixTU3}.}
\begin{table}[t]
  \centering
  \caption{{Driving Performance with Random Traffic Conditions}}
  \resizebox{1.0\linewidth}{!}{
    \begin{tabular}{ccccc}
    \toprule
    Scenario & Travel time (s) & Collision& Rule violation& Succ. rate\\
    \midrule
    ML-ACC     &  $16.97\pm 2.99$  & 3/40 & 0/40  & 92.5\% \\
    Roundabout    &   $20.49\pm 3.26$ & 0/40 &  2/40 & 95\% \\
    Sig-Inter     &   $17.02\pm 2.52$  & 1/40 & 0/40 & 97.5\% \\
    Mixed-Traffic & $24.59\pm 0.21$ & 0/40 & 3/40 & 92.5\%  \\
    \bottomrule
    \end{tabular}%
    }
  \label{tab:robustTest}%
\end{table}%
Nonetheless, the majority of trials demonstrate that the {AV} operates safely and efficiently. For instance, when faced with a slower {surrounding vehicle} ahead, the {AV} can successfully overtake while maintaining an adequate safety distance. Additionally, as an {surrounding vehicle} exits the intersection or roundabout, the {AV} decelerates smoothly to avoid collision and accelerates once the {surrounding vehicle} has moved away. Consequently, this randomly generated simulation of the {surrounding vehicle} demonstrates the effectiveness of the proposed approach in various urban driving situations without the need for hyperparameter modifications.
With the above analysis, we conclude that the proposed UDMC framework is able to generate safe reactions and keep high traffic efficiency with high-level decision-making with flexible maneuvers.

\subsection{{Hazard Analysis}}\label{sec:hazardAnalysis}
{{Following the established standard of ISO~26262 in autonomous vehicle safety~\cite{arab2024safety,gosavi2018application}, }we conduct a detailed hazard analysis of the proposed UDMC framework to evaluate potential risks in practical autonomous driving applications {utilizing Failure Mode and Effects Analysis (FMEA)~\cite{sharma2018failure,arab2024safety}.
To cater to the unique aspects of autonomous driving tasks without the intervention of humans, we tailor the FMEA by omitting the index of the likelihood of detection (D) and modifying the risk priority number to be $\text{RPN}=S\times O$, where $S\in \{s|0\leq s\leq 10\,,s\in \mathbb{N}$ and $O\in \{o|0\leq o\leq 10\,, o\in \mathbb{N}\}$ mean the Severity and Likelihood of Occurrence, respectively.}

{To enhance clarity, we categorize the failure modes of the autonomous driving system into two distinct groups: failures due to the UDMC scheme (first two rows of Table~\ref{tab:hazardAnalysis}) and failures induced by environmental factors (last two rows of Table~\ref{tab:hazardAnalysis}). In terms of the first category, there are mainly two failure modes, wrong prediction of the trajectory of the traffic participants, and V2X uncertainties. In some cases (e.g., merging into a roundabout), the predicted driving direction of the surrounding vehicles is not accurate, leading to possible sideswipes with others. As sideswipes have the potential to induce rollovers of vehicles, severe to life-threatening (survival probable) injuries~\cite{gosavi2018application}, the severity of this failure mode is set as $S=8$. Although its likelihood of occurrence is low, the RPN is relatively high for the high severity of this failure mode.}
{Moreover, it is imperative to recognize the potential risks linked to V2X assumptions regarding transmission stability, low latency, and accurate localization. Therefore, mitigating uncertainties that could emerge during real-world operations is recommended. Given the relatively low likelihood of occurrence for this failure mode, its RPN is assessed as moderate.}

{On the other hand, the simulator does not perfectly simulate the traffic conditions in the real world and the road network is sometimes misleading. First, lane markings should be solid lines when approaching an intersection to prohibit vehicles from changing lanes before the intersection. However, in some scenarios, the above rule for lane markings in the urban map is violated, leading to aggressive lane changing and other possible unsafe behaviors. In the benchmark testing, many similar occasions are detected, which leads to a larger $O$ of this failure mode, leading to the largest RPN in Table~\ref{tab:hazardAnalysis}. Second, the proposed global path searching method finds the nearest path to the destination, which possibly leads to the detection of \textit{deviate from the route} in the benchmark and a lower route completion score in Table~\ref{tab:carlaTown05Benchmark}. As this failure mode only leads to lower traffic efficiency without collision risks, we set $S=2$ and $\text{RPN}=4$ as shown in the last row of Table~\ref{tab:hazardAnalysis}. In summary, this FMEA gives valuable guidance for the further improvement of the proposed autonomous driving system.}

\begin{table}[t]
  \centering
  \caption{{Failure Mode and Effects Analysis of the Proposed UDMC}}
    \begin{tabular}{ccccc}
    \toprule
    Failure Mode & Effect & S     & O     & RPN \\
    \midrule
    Wrong prediction & Crash with other vehicles & 8     & 2     & 16 \\
    V2X uncertainties & Inaccurate localization & 7     & 1     & 7 \\
    \midrule
    Inaccurate lane marks & Aggressive lane changing & 6     & 3     & 18 \\
    False global path & Drive inefficiently & 2     & 2     & 4 \\
    \bottomrule
    \end{tabular}%
  \label{tab:hazardAnalysis}%
\end{table}%

\section{Conclusion}
This work presents a comprehensive framework named UDMC, which integrates traffic object feature extraction by APF, and motion prediction of the traffic participants by IGPR. Collision avoidance and traffic rules compliance are formulated as soft constraints in the OCP with well-designed PFs. Our approach simultaneously achieves high-level decision-making and low-level control, which leads to a computationally efficient pipeline for urban driving. We compare the proposed approach with rule-based methods and a learning-based method named InterFuser on four challenging urban driving scenarios. A series of ablation simulations and benchmark evaluations are conducted as well. The results clearly demonstrate the effectiveness, robustness, and safety of driving behaviors attained by UDMC, and a high success rate and stable commuting time performance are also achieved. As our work provides a general autonomous driving framework, different modules in this framework can be suitably modified to adapt to specific driving applications, owing to the compatibility and generalizability of this framework.
As part of our future work, UDMC can be deployed on equipment with low-configuration industrial computers without plenty of adaptation work.

\bibliographystyle{ieeetr}
\bibliography{refs}

\end{document}